\pdfoutput=1

\documentclass[11pt]{article}

\usepackage[final]{acl}

\usepackage{times}
\usepackage{latexsym}
\usepackage{kotex}
\usepackage{setspace}
\usepackage{multirow}
\usepackage{makecell}
\usepackage{booktabs}
\usepackage{amsmath}
\usepackage{enumitem}
\usepackage{pifont}
\usepackage{subcaption}
\usepackage{color, colortbl}
\usepackage{tcolorbox}
\definecolor{Gray}{gray}{0.9}
\definecolor{My_green}{RGB}{50,205,50}
\definecolor{My_red}{RGB}{204,34,31}

\usepackage[T1]{fontenc}

\usepackage[utf8]{inputenc}

\usepackage{microtype}

\usepackage{inconsolata}

\usepackage{graphicx}

\title{\makecell{\textsc{Picture}: Enhancing Theory-of-Mind in Large Language Models \\ by Revealing, Not Hiding, Characters' Lack of Knowledge}}

\author{Eojin Jeon$^{1}$, \
        SangKeun Lee$^{1,2}$ \\
        $^{1}$Department of Artificial Intelligence, Korea University, Seoul, Republic of Korea  \\
  $^{2}$Department of Computer Science and Engineering, Korea University, Seoul, Republic of Korea \\
  \texttt{\{skdlcm456, yalphy\}@korea.ac.kr} \\}

\begin{document}
\maketitle
\begin{abstract}
Simulating human-like Theory of Mind (ToM) has been a longstanding problem in natural language processing (NLP). To address this, existing works introduce a reasoning step of event hiding (a.k.a. perspective-taking), where events unknown to a character are removed before question answering. However, resorting to event hiding for ToM reasoning presents a performance degradation issue due to the strict output format constraints involved in event hiding. To mitigate this issue, we propose generating perspective-taking outputs as free-form explanations without event hiding, but this poses a notable yet underexplored challenge: LLMs need to inhibit responses to events unknown to characters, because the absence of event hiding exposes LLMs to these events throughout reasoning. To address this challenge, we hypothesize and empirically verify that LLMs can achieve such inhibition if a character’s \textit{lack of knowledge} about events is made explicit during reasoning. Based on this finding, we introduce \textsc{Picture}, a new prompting method that enables LLMs to generate a character's lack of knowledge within free-form CoT. Experimental results show that \textsc{Picture} outperforms existing prompting methods by an average of 7.3\% on false-belief tasks.\footnote{Our code is available at \url{https://github.com/jej127/PICTURE}.}

\end{abstract}

\section{Introduction}
\label{intro}
\begin{figure*}[t!]
\centering
  \includegraphics[width=\textwidth]{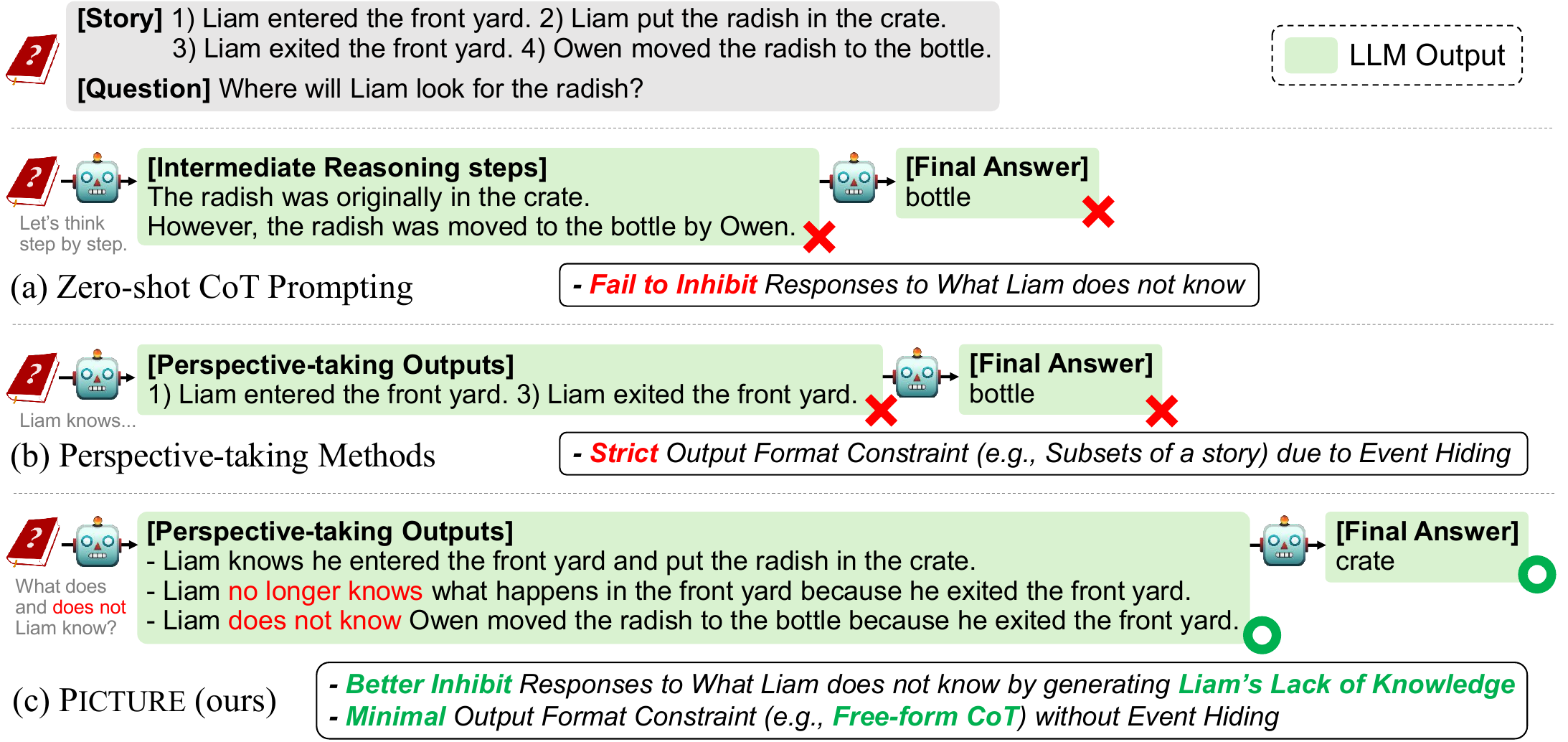}
    \caption{Comparison between (a) zero-shot CoT prompting, (b) the existing perspective-taking methods with event hiding, and (c) our proposed method. Green areas denote LLM outputs.}
    \label{preliminary}
\end{figure*}

Humans utilize a remarkable ability, called Theory of Mind (ToM), to attribute mental states (e.g., beliefs) to others \cite{Baron-1985, Premack-1987}. Despite the advancement of Large Language Models (LLMs), simulating human-like ToM using them has been a longstanding problem in natural language processing (NLP) \cite{Sclar-ACL2023,Wilf-ACL2024,Jung-EMNLP2024}. A key bottleneck of this challenge lies in LLMs' failure to inhibit\footnote{In this work, we use the term inhibition as a behavior-level notion, consistent with psychology literature \cite{Carlson-2001}. Specifically, LLMs are said to inhibit if they correctly answer a ToM question despite the presence of events unknown to the character in the prompt. This definition does not make any claim about inhibition at the level of internal mechanisms (e.g., neurons within LLMs).} responses to events unknown to characters in false-belief tasks. As illustrated in Figure \ref{preliminary}(a), LLMs tend to answer the false-belief question with respect to the true state of reality rather than the character's own perspective \cite{Wilf-ACL2024,Jung-EMNLP2024}.

To address this bottleneck, recent studies \cite{Wilf-ACL2024,Jung-EMNLP2024,Hou-ACL2024,Xu-ACL2025, Sneheel-COLING2025} introduce a preliminary reasoning step called perspective-taking, which is typically implemented via event hiding. In this reasoning step, LLMs are first instructed to remove events unknown to a character from an input story. Then, LLMs answer the question based solely on this filtered story. This allows LLMs to bypass the need to inhibit responses to events unknown to a character during question answering.

While event hiding is a standard practice to implement perspective-taking, resorting to event hiding presents a critical issue that can limit LLMs' ToM capabilities. Specifically, event hiding typically relies on strict output format constraints—such as requiring JSON structures \cite{Jung-EMNLP2024}, graph representations \cite{Xu-ACL2025}, or subsets of the story \cite{Wilf-ACL2024,Hou-ACL2024,Sneheel-COLING2025}—to ensure that events unknown to the character are cleanly separated. However, due to the well-documented trade-off between format adherence and reasoning performance in LLMs \cite{Tam-EMNLP2024, Banerjee-ICML2025}, these format constraints may inadvertently hinder perspective-taking, potentially degrading LLMs' ToM capabilities. This issue is exemplified in Figure \ref{preliminary}(b) where the second event in the story is incorrectly removed during perspective-taking.

To alleviate this issue, instead of relying on event hiding, we propose generating perspective-taking outputs as free-form explanations. By omitting event hiding and by expressing the character’s perspective in a free-form manner, we impose minimal constraints on the output format for perspective-taking. In doing so, we can effectively mitigate the performance degradation stemming from rigid output format constraints. However, eliminating event hiding poses a notable yet underexplored challenge: LLMs need to inhibit responses to events unknown to characters, because the absence of event hiding exposes LLMs to these events throughout reasoning. From a psychology standpoint, addressing this challenge can be understood as eliciting the cognitive ability known as \textit{inhibitory control}, which involves inhibiting responses to irrelevant stimuli while pursuing a specific goal \cite{Carlson-2001}.

To address this challenge, we establish the following hypothesis: if a character’s \textit{lack of knowledge} about events is made explicit during reasoning—for example, through statements such as ``Liam does not know Owen moved the radish to the bottle'' as in Figure \ref{preliminary}(c)—, then LLMs can inhibit responses to those events unknown to the character, using the character's lack of knowledge as hints for inhibition. Empirically, we validate this hypothesis through a preliminary study described in Section \ref{motivation}.

Based on this finding, we propose \textbf{P}erspective-taking w\textbf{i}th Generated La\textbf{c}k of Knowledge in Chain-of-\textbf{T}ho\textbf{u}ght \textbf{Re}asoning (\textsc{Picture}), a new prompting method that instructs LLMs to generate characters' lack of knowledge within free-form Chain-of-Thought (CoT), as illustrated in Figure \ref{preliminary}(c). Specifically, \textsc{Picture} first prompts LLMs to infer what the character does and does not know. To complete the reasoning path that leads to the final answer, we include a trigger sentence ``\textit{Think step by step}'', motivated by \citet{Kojima-NeurIPS2022}.

To show the effectiveness of \textsc{Picture}, we conduct an extensive evaluation on ToM benchmarks. The results demonstrate that \textsc{Picture} achieves a 7.3\% improvement on false-belief tasks compared to existing zero-shot prompting methods. Furthermore, through analysis, we show that LLMs using \textsc{Picture} generate characters' lack of knowledge within free-form CoT as expected.

Our contributions are summarized as follows:

\begin{itemize}
  \item We are the first to show that LLMs can inhibit responses to events unknown to a character for ToM reasoning by making the character's lack of knowledge explicit during reasoning.
  \item We propose \textsc{Picture}, a prompting method to aid LLMs in inhibition for ToM reasoning, which instructs them to generate a character's lack of knowledge within free-form CoT.
  \item Extensive evaluation on ToM benchmarks demonstrates that \textsc{Picture} improves ToM reasoning performance of LLMs over existing zero-shot prompting methods.
\end{itemize}

\begin{figure*}[ht]
\begin{subfigure}{.5\textwidth}
  \centering
  \includegraphics[width=.99\linewidth]{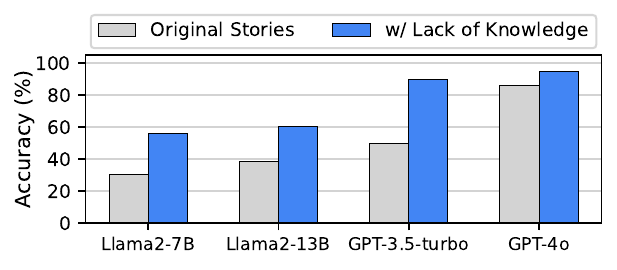}  
  \caption{False Belief}
  \label{fig:pre_a}
\end{subfigure}
\begin{subfigure}{.5\textwidth}
  \centering
  \includegraphics[width=.99\linewidth]{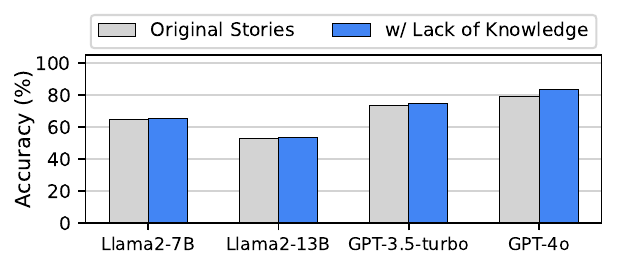}  
  \caption{True Belief}
  \label{fig:pre_b}
\end{subfigure}
\caption{LLM performances on Percept-ToMi. ``w/ Lack of Knowledge'' denotes stories that are augmented with characters' lack of knowledge—i.e., for each event unknown to a character, we prepend a phrase ``\texttt{\{character\} does not know that}'' to that event, explicitly indicating that the event is unknown to the character.}
\label{fig:pre}
\end{figure*}

\section{Related Work}
\subsection{Improving LLMs' ToM with Event Hiding}

Simulating human-like ToM has long been a central challenge in NLP \cite{Sclar-ACL2023,Wilf-ACL2024,Jung-EMNLP2024}. In particular, a crucial bottleneck of this challenge lies in LLMs' failure to inhibit responses to events unknown to characters in false-belief tasks \cite{Wilf-ACL2024,Jung-EMNLP2024}. Consequently, LLMs often answer false-belief questions with respect to the true state of reality rather than a character's own perspective. To address this, a line of work introduces a reasoning step of event hiding (a.k.a. perspective-taking), where events unknown to a character are removed from an input story before question answering. 

Specifically, SimToM \cite{Wilf-ACL2024} and PercepToM \cite{Jung-EMNLP2024} ask LLMs to filter out events using two-step prompting. Other works focus on improving the event hiding results by using external tools or iterative refinement algorithms. For example, SymbolicToM \cite{Sclar-ACL2023} iteratively updates the characters' belief states by representing them as a set of graphs. TimeToM \cite{Hou-ACL2024} and Decompose-ToM \cite{Sneheel-COLING2025} refine characters' beliefs by introducing an external tool called belief solver and by decomposing ToM tasks into recursive simulations, respectively. EnigmaToM \cite{Xu-ACL2025} trains a neural knowledge base to infer the state information for entities. Agentic-ToM \cite{Sarangi-EMNLP2025} employs autonomously invokable cognitive tools to analyze agents’ perspectives.

One notable characteristic of these methods is that LLMs lack access to events unknown to characters during question answering, which is facilitated by event hiding. In doing so, LLMs no longer need to inhibit responses to those events for ToM reasoning.\footnote{From a psychology standpoint, event hiding is conceptually distinct from inhibitory control and therefore can hardly be viewed as an implementation of inhibitory control in LLMs. See Appendix \ref{sec:inhibitory} for more discussion.} In contrast, our method does not rely on event hiding and encourages LLMs to inhibit responses to events unknown to characters, maintaining LLMs' access to those events throughout reasoning.

\subsection{Improving LLMs' ToM without Event Hiding}
A few works attempt to enhance the ToM capability of LLMs without explicitly hiding events. Specifically, DWM \cite{Huang-EMNLP2024} segments the context into multiple state events and updates characters’ beliefs for each state event. ThoughtTracing \cite{Kim-COLM2025} tracks the mental states of characters, following the sequential Monte Carlo structure. Multi-agent ToM prompting \cite{Li-EMNLP2023} incorporates explicit belief state representations for the cooperative text game. However, these methods primarily focus on generating characters’ knowledge, as reflected in their instructions, with limited explicit emphasis on generating characters’ lack of knowledge as intermediate reasoning steps. In contrast, our work aims to generate not only the characters’ knowledge but also their lack of knowledge for effective ToM reasoning.

\begin{figure*}[t!]
\centering
  \includegraphics[width=\textwidth]{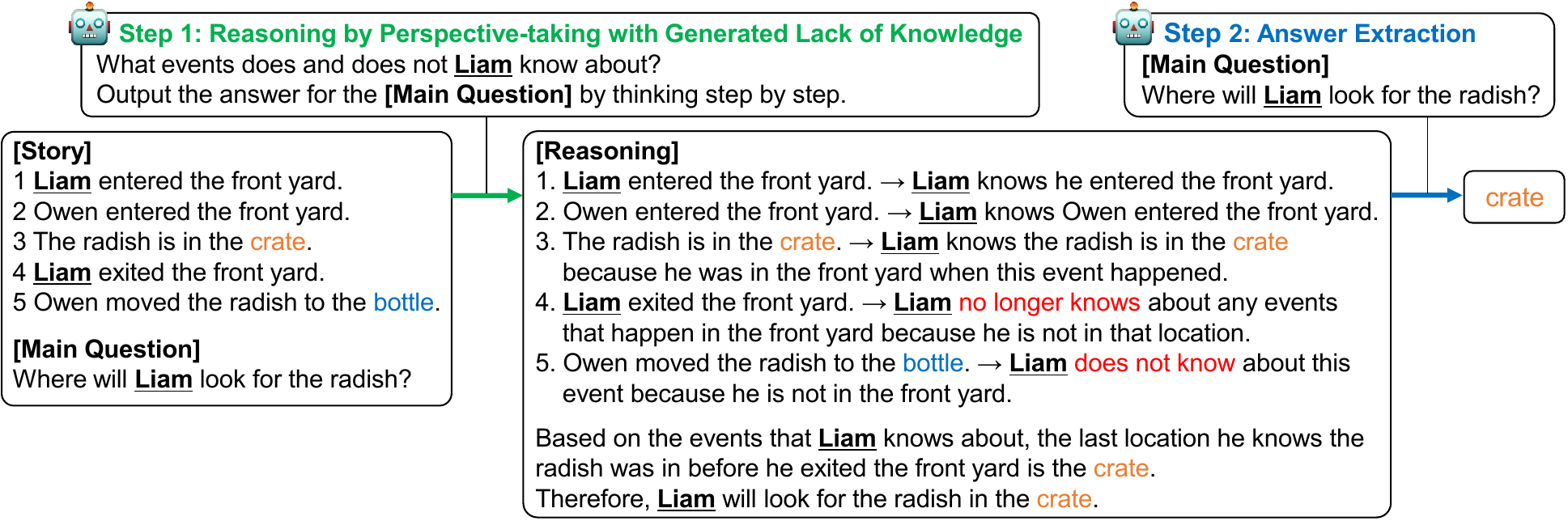}
    \caption{An overview of \textsc{Picture}. In the reasoning by perspective-taking with generated lack of knowledge (Step 1), we first ask LLMs to infer what the character does and does not know within a story. We then include a trigger sentence ``\textit{Think step by step}'', to generate further reasoning steps that lead to the final answer. In the answer extraction step (Step 2), LLMs extract the answer from the reasoning generated in the previous step.}
    \label{fig:pipeline}
\end{figure*}

\section{Preliminary Study}
\label{motivation}

As discussed in Section \ref{intro}, our goal is to develop a prompting method for ToM reasoning that generates perspective-taking outputs as free-form explanations without event hiding. By discarding event hiding and by describing characters' perspectives in a free-form manner, we impose minimal constraints on the output format for perspective-taking. This design choice helps mitigate the performance degradation arising from rigid output format constraints \cite{Tam-EMNLP2024}. However, removing event hiding introduces a remarkable challenge: LLMs need to inhibit responses to events unknown to characters, since the absence of event hiding exposes LLMs to these events throughout reasoning. In this section, we present and empirically validate the hypothesis for addressing this challenge.

To begin with, we formulate the following hypothesis: if a character’s \textit{lack of knowledge} about events is made explicit during reasoning—for instance, through statements such as ``Liam does not know Owen moved the radish to the bottle''—, then LLMs can inhibit responses to those events during reasoning. To verify this hypothesis, we conduct a preliminary study using the Percept-ToMi \cite{Jung-EMNLP2024} dataset. This dataset contains not only the ToM questions and answers, but also the perceiver information for each event (i.e., the information of who perceived the event). To indicate the character's lack of knowledge, we use this perceiver information and append ``\texttt{\{character\} does not know that}'' to each event that the character fails to perceive. We evaluate LLMs separately on the original and modified stories.

The results are presented in Figure \ref{fig:pre}. We observe a substantial improvement in the performance of LLMs on the false-belief task when stories are augmented with the characters' lack of knowledge. In addition, the performances on the true-belief task do not diminish across all LLMs. This suggests that LLMs can effectively inhibit responses to events unknown to characters in false-belief tasks by making the characters' lack of knowledge explicit, thereby confirming our hypothesis. Motivated by this observation, we detail the prompting method that generates characters' lack of knowledge to achieve inhibition in LLMs in Section \ref{method}.

\section{\textsc{Picture}: Prompting to Generate Characters' Lack of Knowledge}
\label{method}

Based on the hypothesis verified in Section \ref{motivation}, we aim to devise a prompting method for ToM reasoning that instructs LLMs to generate characters' lack of knowledge, especially via free-form explanations. To this end, we propose \textsc{Picture} that consists of two steps: 1) reasoning by perspective-taking with generated lack of knowledge (Section \ref{gapt}), and 2) answer extraction (Section \ref{pbi}). Figure \ref{fig:pipeline} illustrates an overview of \textsc{Picture}. The prompts for \textsc{Picture} are available in Appendix \ref{sec:prompts}.

\subsection{Reasoning by Perspective-taking with Generated Lack of Knowledge}
\label{gapt}
To solve a given question with a story, LLMs in this step are expected to 1) provide a free-form CoT that leads to the final answer, and to 2) generate a character's lack of knowledge within that CoT. We adopt CoT as the generation strategy because it is a well-established and effective method for producing free-form reasoning explanations \cite{Chu-ACL2024}. To this end, LLMs are first prompted to infer what the character does and does not know within a story. Different from prior methods which instruct LLMs to hide a character's lack of knowledge (e.g., SimToM), our method instructs LLMs to reveal a character's lack of knowledge during reasoning. To complete the CoT that leads to the final answer, we include a trigger sentence ``\textit{Think step by step}'', inspired by \citet{Kojima-NeurIPS2022}. This facilitates CoT reasoning of LLMs that explicitly involves a character's lack of knowledge.

\subsection{Answer Extraction}
\label{pbi}
In this step, similar to \citet{Kojima-NeurIPS2022}, we prompt LLMs to extract the final answer from the reasoning generated in the previous step. We implement this prompt by asking LLMs to answer the same question based on the generated reasoning. In doing so, LLMs are expected to isolate the final answer for evaluation.

\section{Experimental Settings}

\subsection{Datasets}
\label{setting_dataset}
In the experiment, we primarily aim to evaluate how effectively LLMs can reason about characters' false beliefs using each prompting method. To this end, we use three popular ToM benchmarks that include false-belief questions. Specifically, we use 800 questions from BigToM \cite{Gandhi-NeurIPS2023} and 1,000 questions from ToMi \cite{Le-EMNLP2019} used in \citet{Wilf-ACL2024}. In addition, we use 1,540 belief questions in FANToM \cite{Kim-EMNLP2023}. ToMi consists of structured stories generated based on templates, whereas BigToM features more natural and less structured narratives. FANToM involves multi-party conversations where the specific speaker leaves and rejoins. We use accuracy as the evaluation metric for these benchmarks.

\begin{table*}
\small
\begin{center}
\setlength\tabcolsep{4.8pt}
{
\begin{tabular}{clcccccc}
\toprule
\multirow{2}{*}{Model} & \multirow{2}{*}{Method} & \multicolumn{3}{c}{False Belief} & \multicolumn{3}{c}{All} \\ \cmidrule(lr){3-5}\cmidrule(lr){6-8}
& & BigToM & ToMi & FANToM & BigToM & ToMi & FANToM \\ 
\midrule
 \multirow{5}{*}{Llama2-7B-chat} & Vanilla & $47.3_{\pm1.5}$ & $32.8_{\pm0.9}$ & $15.3_{\pm0.4}$ & $53.6_{\pm1.2}$ & $45.3_{\pm0.8}$ & $42.2_{\pm0.5}$ \\
 & CoT \cite{Kojima-NeurIPS2022} & $42.9_{\pm1.7}$ & $39.2_{\pm0.9}$ & $26.2_{\pm1.0}$ & $55.5_{\pm1.1}$ & $44.8_{\pm1.0}$ & $47.5_{\pm0.7}$ \\
 & PS \cite{Wang-ACL2023} & $44.0_{\pm1.6}$ & $39.1_{\pm1.3}$ & $18.2_{\pm0.3}$ & $50.8_{\pm1.3}$ & $51.4_{\pm0.8}$ & $42.9_{\pm0.5}$ \\
 & SimToM \cite{Wilf-ACL2024} & $70.3_{\pm0.9}$ & $48.2_{\pm0.5}$ & $13.0_{\pm0.8}$ & $57.2_{\pm1.0}$ & $51.8_{\pm0.5}$ & $41.2_{\pm0.7}$ \\ 
 & \textsc{Picture} (Ours) & $\textbf{72.9}_{\pm1.1}$ & $\textbf{53.0}_{\pm1.0}$ & $\textbf{37.8}_{\pm0.8}$ & $\textbf{69.0}_{\pm0.9}$ & $\textbf{59.2}_{\pm0.8}$ & $\textbf{50.3}_{\pm1.0}$ \\ 
 \midrule
\multirow{6}{*}{Llama2-13B-chat} & Vanilla & $40.9_{\pm1.3}$ & $39.3_{\pm1.1}$ & $18.4_{\pm0.5}$ & $52.0_{\pm0.5}$ & $50.9_{\pm1.3}$ & $43.9_{\pm0.3}$ \\
 & CoT \cite{Kojima-NeurIPS2022} & $52.2_{\pm1.1}$ & $31.9_{\pm0.8}$ & $43.4_{\pm1.1}$ & $56.1_{\pm0.8}$ & $46.9_{\pm0.7}$ & $51.5_{\pm1.1}$ \\
 & PS \cite{Wang-ACL2023} & $28.3_{\pm0.6}$ & $38.6_{\pm1.7}$ & $34.2_{\pm0.7}$ & $53.1_{\pm0.2}$ & $43.8_{\pm1.1}$ & $49.1_{\pm0.6}$ \\
 & PercepToM \cite{Jung-EMNLP2024} & $45.2_{\pm1.6}$ & $51.2_{\pm1.2}$ & $38.0_{\pm0.9}$ & $49.0_{\pm0.6}$ & $48.4_{\pm1.0}$ & $49.4_{\pm0.7}$ \\ 
 & SimToM \cite{Wilf-ACL2024} & $61.8_{\pm0.9}$ & $50.6_{\pm0.9}$ & $36.5_{\pm1.0}$ & $58.0_{\pm1.6}$ & $60.8_{\pm1.1}$ & $49.6_{\pm0.8}$ \\ 
 & \textsc{Picture} (Ours) & $\textbf{67.8}_{\pm2.0}$ & $\textbf{64.3}_{\pm1.5}$ & $\textbf{47.7}_{\pm0.8}$ & $\textbf{73.6}_{\pm0.8}$ & $\textbf{64.3}_{\pm0.7}$ & $\textbf{55.5}_{\pm0.2}$ \\ 
 \midrule
\multirow{6}{*}{GPT-3.5-Turbo} & Vanilla & $41.1_{\pm0.9}$ & $64.6_{\pm0.8}$ & $10.8_{\pm0.4}$ & $66.6_{\pm0.7}$ & $68.5_{\pm0.6}$ & $40.9_{\pm0.4}$ \\
 & CoT \cite{Kojima-NeurIPS2022} & $56.1_{\pm0.8}$ & $55.3_{\pm0.9}$ & $49.1_{\pm0.5}$ & $75.9_{\pm0.8}$ & $65.2_{\pm0.7}$ & $59.0_{\pm0.5}$ \\
 & PS \cite{Wang-ACL2023} & $55.6_{\pm0.5}$ & $59.2_{\pm0.6}$ & $44.9_{\pm0.7}$ & $71.6_{\pm0.8}$ & $66.3_{\pm0.6}$ & $56.1_{\pm1.0}$ \\
 & PercepToM \cite{Jung-EMNLP2024} & $45.7_{\pm1.1}$ & $79.7_{\pm0.8}$ & $25.3_{\pm0.4}$ & $68.4_{\pm0.3}$ & $70.5_{\pm0.5}$ & $50.8_{\pm0.6}$ \\ 
 & SimToM \cite{Wilf-ACL2024} & $70.4_{\pm0.6}$ & $81.0_{\pm0.6}$ & $27.0_{\pm0.6}$ & $81.5_{\pm0.5}$ & $72.6_{\pm0.4}$ & $49.9_{\pm0.9}$ \\ 
 & \textsc{Picture} (Ours) & $\textbf{86.0}_{\pm0.5}$ & $\textbf{83.5}_{\pm0.9}$ & $\textbf{58.1}_{\pm0.8}$ & $\textbf{88.7}_{\pm0.6}$ & $\textbf{77.5}_{\pm0.6}$ & $\textbf{65.0}_{\pm1.0}$ \\ 
 \midrule
 \multirow{6}{*}{GPT-4o} & Vanilla & $88.6_{\pm0.3}$ & $80.2_{\pm0.6}$ & $52.5_{\pm0.5}$ & $92.2_{\pm0.4}$ & $74.6_{\pm0.7}$ & $65.4_{\pm0.4}$ \\
 & CoT \cite{Kojima-NeurIPS2022} & $90.2_{\pm0.4}$ & $83.3_{\pm0.8}$ & $76.8_{\pm0.6}$ & $93.9_{\pm0.3}$ & $78.9_{\pm0.5}$ & $79.9_{\pm0.5}$ \\
 & PS \cite{Wang-ACL2023} & $85.9_{\pm0.2}$ & $89.1_{\pm0.4}$ & $76.4_{\pm0.4}$ & $91.3_{\pm0.3}$ & $77.3_{\pm0.4}$ & $78.9_{\pm1.0}$ \\
 & PercepToM \cite{Jung-EMNLP2024} & $87.5_{\pm0.3}$ & $88.1_{\pm0.3}$ & $79.4_{\pm0.6}$ & $87.4_{\pm0.3}$ & $76.3_{\pm0.4}$ & $84.1_{\pm0.6}$ \\ 
 & SimToM \cite{Wilf-ACL2024} & $91.5_{\pm0.3}$ & $88.1_{\pm0.4}$ & $79.6_{\pm0.7}$ & $91.9_{\pm0.3}$ & $79.4_{\pm0.4}$ & $81.3_{\pm0.9}$ \\ 
 & \textsc{Picture} (Ours) & $\textbf{96.5}_{\pm0.6}$ & $\textbf{90.5}_{\pm0.2}$ & $\textbf{91.5}_{\pm0.7}$ & $\textbf{97.3}_{\pm0.4}$ & $\textbf{91.8}_{\pm0.5}$ & $\textbf{87.3}_{\pm1.0}$ \\ 
 \bottomrule
\end{tabular}
}
\end{center}
\caption{Results on the BigToM, ToMi, and FANToM datasets. Accuracy is reported as the evaluation metric, with subscripts indicating standard deviations. The best-performing method is shown in \textbf{bold}. All improvements achieved by \textsc{Picture} are statistically significant according to a t-test with $p < 0.05$.}
\label{tab:main}
\end{table*}

\subsection{Models}
Similar to \citet{Wilf-ACL2024}, we evaluate each method using four widely used LLMs. Among the open-source LLMs, we employ Llama2-7B-chat and Llama2-13B-chat \cite{Touvron-Corr2023}. For the closed-source models, we use GPT-3.5-Turbo \cite{OpenAI35} and GPT-4o \cite{OpenAI4o}. We perform inference with the open-source models using Hugging Face and access the closed-source models through the OpenAI API.\footnote{\url{https://platform.openai.com/docs/api-reference/}} We set the temperature of $\tau=0.0$ and top-$p$ of 1.0 for the closed-source models.

\subsection{Baselines}
We compare \textsc{Picture} with two categories of zero-shot prompting methods: (1) methods to improve general reasoning capabilities, and (2) methods dedicated to improving ToM capabilities. Prompts for each method are provided in Appendix \ref{sec:prompts_base}.

The methods in the first category have demonstrated potential across a wide range of reasoning tasks, such as arithmetic and commonsense reasoning. These methods include: (1) Vanilla prompting, where the LLM is asked to select an answer without rationales; (2) CoT prompting \cite{Kojima-NeurIPS2022}, which decomposes a problem into multiple intermediate steps to be solved; and (3) Plan-and-Solve (PS) prompting \cite{Wang-ACL2023}, which generates a plan that divides the task into subtasks. For Vanilla and CoT, we use the prompts used in \citet{Wilf-ACL2024}. For PS, we use the prompts according to their published settings.


The methods in the second category are dedicated to improving ToM capabilities. They either hide events unknown to characters, or iteratively track characters' knowledge without event hiding. We consider methods that, like ours, do not rely on external tools or iterative refinement algorithms for reasoning. These methods include: SimToM \cite{Wilf-ACL2024} and PercepToM \cite{Jung-EMNLP2024}, both of which rely on event hiding. We use the prompts as described in their respective papers.


\begin{table*}[h]
\small
\centering
\setlength\tabcolsep{3.0pt}
{
\begin{tabular}{p{15.5cm}}
\toprule

\rowcolor{blue!20}
\multicolumn{1}{c}{Example 1 (BigToM)}  \\
\midrule
\textbf{Context}: Amaury is a baker and wants to make a pastry for the customers visiting his bakery. Amaury takes a jar of raisins from the pantry to add to the pastry dough. His assistant, while tidying the pantry, accidentally replaces the raisin jar with a jar containing sultanas without realizing the difference. Amaury does not notice the jar swap.\\

\textbf{Question}: Does Amaury believe the jar contains a) raisins or b) sultanas? \\

\midrule
\rowcolor[gray]{.9}
 SimToM (baseline) \\
\textbf{Perspective-taking Outputs}: \\
\; Amaury is a baker and wants to make a pastry for the customers visiting his bakery. Amaury takes a jar of raisins from \\
\; the pantry to add to the pastry dough. \underline{His assistant, while tidying the pantry, accidentally replaces the raisin jar with a jar} \\
\; \underline{containing sultanas without realizing the difference.} \\
\textbf{Answer}: Amaury believes the jar contains b) sultanas. (\textcolor{My_red}{\ding{55}}) \\

\midrule
\rowcolor[gray]{.9}
\textsc{Picture} (ours) \\
\textbf{Perspective-taking Outputs}: \\
\; Amaury knows that he took a jar of what he believed to be raisins from the pantry to add to the pastry dough. He is also \\
\; aware that his assistant accidentally replaced the raisin jar with a jar containing sultanas without realizing the difference. \\
\; However, the story mentions that Amaury did not notice the jar swap. Given this information, \textcolor{My_red}{Amaury is not aware of the} \\
\; \textcolor{My_red}{assistant's mistake in replacing the raisins with sultanas.} \\

\textbf{Answer}: Amaury believes the jar contains a) raisins. (\textcolor{My_green}{\ding{51}}) \\
\midrule
\rowcolor{blue!20}
\multicolumn{1}{c}{Example 2 (ToMi)} \\ 
\midrule
\textbf{Context}: 1 Liam entered the front yard. 2 Jackson entered the front yard. 3 Owen entered the front yard. 4 The radish is in the crate. 5 Liam exited the front yard. 6 Owen moved the radish to the bottle. \\

\textbf{Question}: Where will Liam look for the radish? Choose from the following: crate, bottle \\
\midrule
\rowcolor[gray]{.9}
SimToM (baseline) \\
\textbf{Perspective-taking Outputs}: \\ 
\; 1 Liam entered the front yard. 5 Liam exited the front yard. \\
\textbf{Answer}: Liam will look for the radish in the bottle. (\textcolor{My_red}{\ding{55}}) \\
\midrule

\rowcolor[gray]{.9}
 \textsc{Picture} (ours) \\
\textbf{Perspective-taking Outputs}: \\
\; Let's break down the events step by step to determine what Liam knows: \\
\; 1. Liam entered the front yard. - Liam knows that he entered the front yard. \\

\; 2. Jackson entered the front yard. - Liam knows that Jackson entered the front yard. \\

\; 3. Owen entered the front yard. - Liam knows that Owen entered the front yard. \\

\; 4. The radish is in the crate. - Liam knows that the radish is in the crate. \\

\; 5. Liam exited the front yard. - \textcolor{My_red}{Liam no longer knows about any events that happen in the front yard.} \\

\; 6. Owen moved the radish to the bottle. - \textcolor{My_red}{Liam does not know about this event.} \\

\; Based on what Liam knows about, the last location he knows the radish was in before he exited the front yard is the crate. \\
\textbf{Answer}: Liam will look for the radish in the crate. 
 (\textcolor{My_green}{\ding{51}}) \\

 \bottomrule
\end{tabular}
}
\caption{Qualitative examples comparing SimToM and \textsc{Picture} on false-belief questions from the BigToM and ToMi datasets. \textcolor{My_red}{Red} text indicates the character's lack of knowledge generated by \textsc{Picture}. \underline{Underlined} text highlights events unknown to the character but are incorrectly retained by SimToM after perspective-taking.}
\label{tab:qual}
\end{table*}

\section{Results}
\label{exp_result}
In this section, we pose several key research questions (RQs) and answer each through experiments.
\begin{itemize}[noitemsep,leftmargin=*]
\item \textbf{RQ1}: Do LLMs better inhibit responses to events unknown to characters using \textsc{Picture}? (§\ref{main})
\item \textbf{RQ2}: Does \textsc{Picture} enable LLMs to perform perspective-taking better than baselines? (§\ref{qual})
\item \textbf{RQ3}: Does \textsc{Picture} encourage LLMs to generate characters' lack of knowledge within free-form CoT as expected? (§\ref{qual})
\item \textbf{RQ4}: Can \textsc{Picture} generalize to different types of ToM questions? (§\ref{opentom})
\item \textbf{RQ5}: Do existing methods based on event hiding perform comparably to \textsc{Picture} when additionally using CoT prompts? (§\ref{ablation})
\end{itemize}

\subsection{Main Results}
\label{main}
The performances of \textsc{Picture} and the baselines are presented in Table \ref{tab:main}. Following \citet{Wilf-ACL2024}, we report the results on the false-belief questions (False Belief) and those on the full set of questions (All). The results are averaged over four runs with random seeds \{0, 111, 222, 333\}. We do not report the performance of the Llama2-7B-chat model using PercepToM, as this model repetitively fails to generate a valid JSON array for over 95\% of the questions during the perception inference stage of PercepToM.

\textsc{Picture} outperforms all baselines across datasets and backbone LLMs. In particular, it achieves an average improvement of 7.3\% on false-belief questions. While prior work has shown that LLMs without explicit perspective-taking prompts (e.g., CoT) often struggle to inhibit responses to events unknown to characters \cite{Wilf-ACL2024}, we demonstrate that \textsc{Picture} mitigates this issue and enhances inhibition in LLMs, outperforming all baselines that do not involve explicit perspective-taking prompts (i.e., Vanilla, CoT, and PS).

Furthermore, \textsc{Picture} surpasses perspective-taking methods based on event hiding, such as SimToM and PercepToM. Although event hiding allows LLMs to circumvent the need for inhibition during question answering, we show that moving beyond this paradigm is viable. In particular, LLMs can inhibit responses to events unknown to characters via \textsc{Picture}. Overall, LLMs using \textsc{Picture} better inhibit responses to events unknown to characters than baselines, showing performance improvements over them especially on false-belief tasks. A detailed breakdown of the results across different question types can be found in Appendix \ref{breakdown}. In addition, we present additional results using more recent LLMs in Appendix \ref{quan_llm}, and provide a comparison between \textsc{Picture} and more recent event hiding methods in Appendix \ref{comp_recent_baseline}.


\begin{table*}
\small
\begin{center}
\setlength\tabcolsep{6.4pt}
{
\begin{tabular}{lccccccc}
\toprule
Method & $\text{Loc}_{c}(\text{F})$ & $\text{Loc}_{c}(\text{S})$ & $\text{Loc}_{f}(\text{F})$ & $\text{Loc}_{f}(\text{S})$ & $\text{MHop}(\text{F})$ & $\text{MHop}(\text{S})$ & Att \\
\midrule
Vanilla & $43.9_{\pm2.1}$ & $32.3_{\pm2.3}$ & $51.5_{\pm1.1}$ & $28.6_{\pm0.5}$ & $46.8_{\pm1.0}$ & $33.4_{\pm0.8}$ & $41.0_{\pm1.2}$ \\
CoT \cite{Kojima-NeurIPS2022} & $59.1_{\pm1.5}$ & $44.0_{\pm2.8}$ & $51.2_{\pm0.9}$ & $28.7_{\pm0.4}$ & $52.5_{\pm0.9}$ & $42.5_{\pm1.0}$ & $46.8_{\pm1.5}$ \\
PS \cite{Wang-ACL2023} & $63.3_{\pm1.8}$ & $54.0_{\pm2.0}$ & $65.7_{\pm1.5}$ & $33.5_{\pm0.6}$ & $64.3_{\pm1.5}$ & $46.6_{\pm0.8}$ & $49.3_{\pm1.3}$ \\
PercepToM \cite{Jung-EMNLP2024} & $54.9_{\pm1.7}$ & $35.1_{\pm2.1}$ & $57.3_{\pm1.2}$ & $33.6_{\pm0.7}$ & $47.6_{\pm1.0}$ & $35.5_{\pm0.7}$ & $39.1_{\pm1.0}$ \\ 
SimToM \cite{Wilf-ACL2024} & $62.5_{\pm1.4}$ & $45.8_{\pm2.4}$ & $53.3_{\pm1.3}$ & $28.7_{\pm0.5}$ & $53.9_{\pm1.3}$ & $34.4_{\pm0.8}$ & $41.6_{\pm1.4}$ \\ 
\textsc{Picture} (Ours) & $\textbf{85.4}_{\pm1.6}$ & $\textbf{62.0}_{\pm1.8}$ & $\textbf{77.1}_{\pm1.4}$ & $\textbf{43.9}_{\pm0.4}$ & $\textbf{68.0}_{\pm1.2}$ & $\textbf{48.0}_{\pm0.5}$ & $\textbf{52.1}_{\pm1.5}$ \\ 
 \bottomrule
\end{tabular}
}
\end{center}
\caption{Results on the OpenToM dataset, which includes questions about object locations (Loc), multi-hop reasoning (MHop), and characters' attitudes (Att). We report F1 score as the evaluation metric following \citet{Xu-ACL2024}, with subscripts indicating standard deviations. The best-performing method is shown in \textbf{bold}.}
\label{tab:opentom}
\end{table*}

\begin{table*}
\small
\begin{center}
\setlength\tabcolsep{9.2pt}
{
\begin{tabular}{lcccccc}
\toprule
\multirow{2}{*}{Method} & \multicolumn{3}{c}{False Belief} & \multicolumn{3}{c}{All} \\ \cmidrule(lr){2-4}\cmidrule(lr){5-7}
& BigToM & ToMi & FANToM & BigToM & ToMi & FANToM \\ 
\midrule
CoT \cite{Kojima-NeurIPS2022} & $56.1_{\pm0.8}$ & $55.3_{\pm0.9}$ & $49.1_{\pm0.5}$ & $75.9_{\pm0.8}$ & $65.2_{\pm0.7}$ & $59.0_{\pm0.5}$ \\
SimToM \cite{Wilf-ACL2024} & $70.4_{\pm0.6}$ & $81.0_{\pm0.6}$ & $27.0_{\pm0.6}$ & $81.5_{\pm0.5}$ & $72.6_{\pm0.4}$ & $49.9_{\pm0.9}$ \\ 
SimToM + CoT & $74.2_{\pm0.3}$ & $81.5_{\pm0.5}$ & $52.8_{\pm1.0}$ & $82.4_{\pm0.7}$ & $73.0_{\pm0.9}$ & $58.8_{\pm0.7}$ \\
\textsc{Picture} (Ours) & $\textbf{86.0}_{\pm0.5}$ & $\textbf{83.5}_{\pm0.9}$ & $\textbf{58.1}_{\pm0.8}$ & $\textbf{88.7}_{\pm0.6}$ & $\textbf{77.5}_{\pm0.6}$ & $\textbf{65.0}_{\pm1.0}$ \\ 
 \bottomrule
\end{tabular}
}
\end{center}
\caption{Ablation study results for \textsc{Picture}. Accuracy is reported as the evaluation metric, with subscripts indicating standard deviations. The best-performing method is shown in \textbf{bold}.}
\label{tab:ablation}
\end{table*}

\subsection{Case Study}
\label{qual}
To better understand why \textsc{Picture} enhances the ToM capabilities of LLMs, we manually inspect the perspective-taking outputs generated by GPT-3.5-Turbo using SimToM and \textsc{Picture}. Table \ref{tab:qual} presents examples of the results on false-belief questions from the BigToM and ToMi datasets. For both examples, we observe that the LLM using SimToM makes errors in perspective-taking, either by incorrectly retaining events unknown to the character (upper), or by falsely removing events known to the character (lower). In contrast, the LLM using \textsc{Picture} provides the correct perspective-taking results, generating the characters' lack of knowledge within free-form CoTs.

To further investigate these reasoning behaviors, we sample and analyze the results from 50 false-belief questions in ToMi. We observe that the LLM using SimToM misclassifies events unknown to the character as known for 18\% of the questions, whereas the LLM using \textsc{Picture} makes the same type of error for only 2\% of the questions. Moreover, the LLM using SimToM misclassifies events known to the character as unknown for 30\% of the questions, whereas the LLM using \textsc{Picture} makes the same type of error for only 6\% of the questions. Lastly, for all 50 questions, we observe that the LLM using \textsc{Picture} generates the characters' lack of knowledge within free-form CoTs.


These results suggest that LLMs can perform perspective-taking better and more robustly through \textsc{Picture} compared to baselines, which in turn enhances subsequent ToM reasoning. Notably, \textsc{Picture} achieves these gains by expressing perspective-taking outputs as free-form explanations. This observation aligns with prior work by \citet{Tam-EMNLP2024}, which shows that relaxing output format constraints can improve LLM reasoning performance. Lastly, \textsc{Picture} helps LLMs to generate characters' lack of knowledge within free-form CoT as expected. More qualitative results using different datasets and LLMs can be found in Appendix \ref{qual_add}. In addition, we present the error analysis of \textsc{Picture} in Appendix \ref{qual_error}.

\subsection{Generalization to Different types of ToM Questions}
\label{opentom}
While the benchmarks considered in Section \ref{main} are representative ToM benchmarks, these benchmarks primarily focus on characters' mental states associated with the physical world (e.g., the location of an object). In contrast, they place less emphasis on mental states associated with the psychological world (e.g., characters' attitudes towards situations). To verify the effectiveness of \textsc{Picture} across different types of questions, we evaluate our method on OpenToM \cite{Xu-ACL2024}, a comprehensive benchmark that covers questions involving both the physical and psychological worlds.

Specifically, OpenToM includes questions about object locations (Loc), multi-hop reasoning (MHop), and characters' attitudes (Att). Questions about object locations and multi-hop reasoning are further divided into first-order (F) and second-order (S) questions. Additionally, questions about object locations are further categorized into coarse- and fine-grained levels, denoted by the subscripts $c$ and $f$, respectively. We use GPT-3.5-Turbo as the backbone model and show the results in Table \ref{tab:opentom}.

\textsc{Picture} outperforms the baselines across all question types. Notably, significant performance gains are observed on questions related to object locations (Loc). \textsc{Picture} is specifically designed to help LLMs explicitly represent a character’s knowledge state (i.e., what the character knows or does not know), to reduce errors arising from incorrectly using information unknown to the character. This design is most directly aligned with location-based questions, where performance critically depends on suppressing information unknown to the character. Consequently, the largest gains are observed in these tasks.

In contrast, for questions in multi-hop reasoning (MHop) and attitude (Att), the performance gain over the second-best method, PS, is relatively smaller. For MHop, this type of question requires additional compositional reasoning beyond knowledge state tracking. While the correct representation of knowledge states remains necessary, it is not sufficient for solving the task, which likely reduces the relative contribution of the inhibitory control trigger. Moreover, the questions in Att often assume that the character has access to all relevant events (e.g., "What would be Sam's attitude towards Amy's action, assuming he observed it?"). In such cases, epistemic conflict is weaker, and knowledge state reasoning plays a smaller role. Nonetheless, our method consistently outperforms these baselines by explicitly reasoning about what the character knows—or does not know—before answering the question. Additional results on ToM questions of more diverse types are available in Appendix \ref{quan_bench}.

\subsection{Ablation Study}
\label{ablation}

While \textsc{Picture} involves CoT prompts (e.g., ``\textit{Think step by step}'') to elicit free-form reasoning explanations, we are curious about whether performance comparable to \textsc{Picture} can be achieved by simply adding CoT prompts to existing methods based on event hiding. To investigate this, we introduce another baseline, SimToM + CoT, wherein 1) events unknown to a character are first removed as in SimToM, and then 2) LLMs perform CoT reasoning based solely on events known to the character. Notably, unlike \textsc{Picture}, this baseline relies on event hiding and does not generate characters’ lack of knowledge, while other components such as CoT prompts are held constant across the two methods.

As shown in Table \ref{tab:ablation}, SimToM + CoT does not yield consistent performance gains over SimToM and CoT prompting. This suggests that naively adding CoT prompts to existing perspective-taking methods provides limited and unstable benefit for ToM reasoning. In contrast, \textsc{Picture} significantly outperforms SimToM + CoT, highlighting that the advantage of \textsc{Picture} over SimToM is not attributed solely to the use of CoT prompts. Rather, its strength stems from omitting event hiding and thereby enabling LLMs to generate characters’ lack of knowledge. Additional results for further variants of \textsc{Picture} are in Appendix \ref{abl_add}.


\section{Conclusion}
In this paper, we explore how LLMs can inhibit responses to events unknown to characters for effective ToM reasoning. To this end, we first hypothesize and empirically verify that LLMs can achieve such inhibition if a character’s \textit{lack of knowledge} about events is made explicit during reasoning. Based on this finding, we propose \textsc{Picture}, a new prompting method that enables LLMs to generate characters' lack of knowledge within free-form CoT. Experimental results show that \textsc{Picture} outperforms existing prompting methods on ToM reasoning tasks, suggesting that LLMs using \textsc{Picture} can inhibit responses to events unknown to characters. Furthermore, analysis verifies that \textsc{Picture} helps LLMs to generate characters' lack of knowledge within free-form CoT as expected.

\section*{Limitations}
While our method has demonstrated effectiveness in ToM reasoning of LLMs, it has three main limitations that should be addressed in future work.

(1) Following the direction of prior work aimed at improving LLMs' ToM capabilities, we primarily focused on text-based ToM benchmarks. However, some recent benchmarks, such as MMToM-QA \cite{Jin-ACL2024}, evaluate ToM in multimodal scenarios involving both text and video. Handling multimodal scenarios introduces additional challenges, such as grounding the meaning of natural language sentences in visual data. Therefore, extending \textsc{Picture} to multimodal ToM remains an important area for future exploration.

(2) Currently, we implement \textsc{Picture} such that LLMs perform perspective-taking within a single step of CoT reasoning. While our method shows the effectiveness across various ToM reasoning tasks, incorporating inference-time algorithms that involve advanced components such as complexity-aware decomposition \cite{Huang-EMNLP2024} or hypothesis generation \cite{Kim-COLM2025} may provide further benefit in modeling false beliefs of characters. Thus, we leave extensions of \textsc{Picture} in this direction to future work.

(3) \textsc{Picture} critically relies on correctly generated lack of knowledge, and errors in this intermediate step could affect downstream reasoning. While our analysis over 50 sampled questions in Section \ref{qual} provides empirical evidence regarding the robustness of this intermediate step, we acknowledge that this analysis relies on a manual inspection of a sampled subset. Extending robustness evaluation to a fully automated large-scale setting is therefore an important avenue for future work.

\section*{Ethical Considerations}
We acknowledge that enhancing LLMs’ ability to reason about others’ beliefs and knowledge states may introduce societal risks if deployed irresponsibly. For instance, improved false-belief reasoning could be misused in deceptive dialog systems, in persuasive chatbots that exploit incorrect user beliefs, or in systems that attempt to infer mental states without consent. Such capabilities may also increase risks of manipulation in politically or financially motivated contexts. 

In addition, benchmarking ToM reasoning on narrative datasets raises concerns about demographic bias, as stories may encode implicit assumptions about different social groups. If not carefully evaluated, improved ToM performance could inadvertently reinforce or amplify such biases in downstream applications. At the same time, perspective-aware reasoning may enable more empathetic and socially sensitive AI systems. Our experiments focus on offline benchmark evaluation rather than real-world deployment. We emphasize that future research involving interactive or user-facing systems should carefully assess potential misuse scenarios, bias amplification, and unintended social consequences.

\section*{Acknowledgments}
This work was supported by the National Research Foundation of Korea (NRF) grant funded by the Korea government (MSIT) (No.RS-2025-00517221 and No.RS-2024-00415812) and Institute of Information \& communications Technology Planning \& Evaluation (IITP) grant funded by the Korea government (MSIT) (No.RS-2024-00439328, Karma: Towards Knowledge Augmentation for Complex Reasoning (SW Starlab), No.RS-2024-00457882, AI Research Hub Project, and No.RS-2019-II190079, Artificial Intelligence Graduate School Program (Korea University)).

\bibliography{custom}

\clearpage

\appendix

\section{Distinction between Event Hiding and Inhibitory Control}
\label{sec:inhibitory}

In psychology, inhibitory control refers to the ability to inhibit responses to irrelevant stimuli while pursuing a cognitively represented goal \cite{Carlson-2001}. As one of the core executive functions \cite{Diamond-2013}, inhibitory control is closely associated with human ToM development \cite{Carlson-2001, Carlson-2002}. Existing methods for ToM reasoning typically implement perspective-taking through event hiding, wherein events unknown to a character are removed from the input prior to question answering. However, we argue that event hiding is conceptually distinct from inhibitory control and therefore can hardly be viewed as an implementation of inhibitory control in LLMs.

First, according to the aforementioned definition of inhibitory control, irrelevant stimuli themselves are not eliminated for inhibition; rather, it is a response to those stimuli that is suppressed when inhibitory control operates. In contrast, event hiding prompts LLMs to perform perspective-taking by eliminating irrelevant stimuli altogether—namely, events unknown to the character. Therefore, this process of elimination is fundamentally misaligned with the core concept of inhibitory control.

\begin{figure}
\begin{center}
\includegraphics[width=1.0\linewidth]{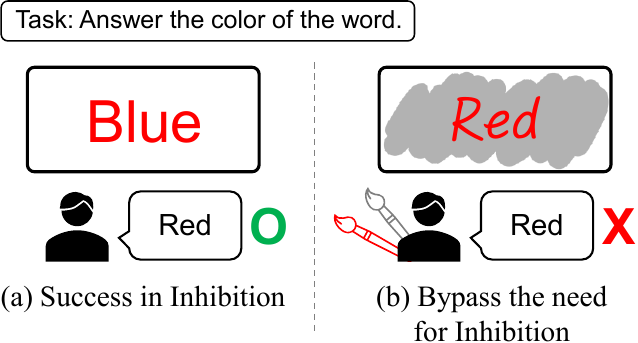}
\caption{Comparison between (a) a correct response in the Stroop test and (b) an impermissible simplification of the test where the original color word ``Blue'' is removed by the test participant. Existing methods based on event hiding are analogous to case (b), since these methods attempt to remove task-irrelevant stimuli (i.e., events unknown to a character) before answering the question.}
\label{comparison}
\end{center}
\end{figure}

Moreover, event hiding differs in principle from how inhibitory control is operationalized in standard psychological assessments such as the Stroop test \cite{Stroop-1935} or the day-night test \cite{Gerstadt-1994}. In the Stroop test, for example, participants are presented with color words printed in incongruent ink colors (e.g., ``\textcolor{red}{blue}''). Their task is to report the ink color (e.g., red) rather than reading the word itself (e.g., blue), as in Figure \ref{comparison}(a). Crucially, task-irrelevant stimuli—such as the meaning of words—remain present and unaltered, and participants are not allowed to remove or modify the stimuli to simplify the task (e.g., by rewriting the word as ``\textcolor{red}{red},'' as in Figure \ref{comparison}(b)). In this sense, event hiding is analogous to such impermissible simplification of this classic test, since it attempts to remove task-irrelevant stimuli, i.e., events unknown to the character, before answering the question.

In summary, event hiding is fundamentally different from inhibitory control because it seeks to eliminate irrelevant stimuli altogether rather than requiring the inhibition of responses to them. Consequently, while event hiding may be an effective strategy for ToM reasoning, it circumvents the need for inhibitory control rather than implementing it. Motivated by this distinction, we aim to implement inhibitory control in LLMs directly, rather than relying on methods that avoid the need for inhibition through event hiding.

\section{Additional Experiments}
\label{sec:add_exp}

\subsection{Results on More Recent LLMs}
\label{quan_llm}
To further demonstrate the effectiveness of \textsc{Picture} using more recent LLMs, we employ Llama3.1-8B-Instruct \cite{Dubey-2024}, Qwen3-8B \cite{Yang-2025}, and Gemma3-12B-Instruct \cite{Gemma-2025} for additional evaluations. We use the same benchmarks as in Section \ref{setting_dataset} and show the results in Table \ref{tab:main_addllm}. Overall, \textsc{Picture} achieves consistent improvements over the competitive methods across all three LLMs. This further indicates the efficacy of \textsc{Picture} on even more recent LLMs, similar to the results in Section \ref{exp_result}.

\begin{table*}
\small
\begin{center}
\setlength\tabcolsep{4.8pt}
{
\begin{tabular}{clcccccc}
\toprule
\multirow{2}{*}{Model} & \multirow{2}{*}{Method} & \multicolumn{3}{c}{False Belief} & \multicolumn{3}{c}{All} \\ \cmidrule(lr){3-5}\cmidrule(lr){6-8}
& & BigToM & ToMi & FANToM & BigToM & ToMi & FANToM \\ 
\midrule
 \multirow{6}{*}{Llama3.1-8B-Inst.} & Vanilla & $58.8_{\pm1.6}$ & $54.7_{\pm1.3}$ & $41.5_{\pm0.4}$ & $68.1_{\pm1.1}$ & $70.4_{\pm1.4}$ & $53.4_{\pm0.3}$ \\
 & CoT \cite{Kojima-NeurIPS2022} & $61.8_{\pm1.5}$ & $53.0_{\pm1.2}$ & $52.7_{\pm0.8}$ & $73.9_{\pm1.3}$ & $63.9_{\pm1.1}$ & $55.3_{\pm0.5}$ \\
 & PS \cite{Wang-ACL2023} & $65.3_{\pm1.7}$ & $62.1_{\pm1.0}$ & $49.9_{\pm0.5}$ & $73.3_{\pm1.1}$ & $65.1_{\pm0.9}$ & $57.6_{\pm0.4}$ \\
 & PercepToM \cite{Jung-EMNLP2024} & $67.3_{\pm1.2}$ & $72.4_{\pm1.4}$ & $53.0_{\pm0.7}$ & $70.8_{\pm0.5}$ & $74.8_{\pm1.3}$ & $57.7_{\pm0.6}$ \\ 
 & SimToM \cite{Wilf-ACL2024} & $73.9_{\pm0.7}$ & $69.5_{\pm1.1}$ & $53.4_{\pm0.7}$ & $62.5_{\pm1.0}$ & $76.4_{\pm0.6}$ & $59.3_{\pm0.5}$ \\ 
 & \textsc{Picture} (Ours) & $\textbf{76.5}_{\pm1.0}$ & $\textbf{82.3}_{\pm1.2}$ & $\textbf{55.5}_{\pm0.6}$ & $\textbf{79.8}_{\pm0.8}$ & $\textbf{79.6}_{\pm0.7}$ & $\textbf{61.4}_{\pm0.8}$ \\ 
 \midrule
\multirow{6}{*}{Qwen3-8B} & Vanilla & $61.3_{\pm1.0}$ & $62.3_{\pm1.2}$ & $28.7_{\pm0.7}$ & $69.8_{\pm0.6}$ & $67.1_{\pm1.1}$ & $50.7_{\pm0.5}$ \\
 & CoT \cite{Kojima-NeurIPS2022} & $73.3_{\pm1.4}$ & $72.0_{\pm1.2}$ & $59.9_{\pm1.0}$ & $77.6_{\pm0.6}$ & $70.7_{\pm0.9}$ & $66.5_{\pm1.3}$ \\
 & PS \cite{Wang-ACL2023} & $82.0_{\pm0.8}$ & $69.5_{\pm1.3}$ & $65.0_{\pm0.5}$ & $82.8_{\pm0.8}$ & $68.2_{\pm1.4}$ & $66.9_{\pm0.9}$ \\
 & PercepToM \cite{Jung-EMNLP2024} & $63.0_{\pm1.4}$ & $76.8_{\pm1.0}$ & $57.9_{\pm0.8}$ & $66.4_{\pm0.8}$ & $77.4_{\pm1.3}$ & $65.9_{\pm0.9}$ \\ 
 & SimToM \cite{Wilf-ACL2024} & $74.3_{\pm1.1}$ & $75.0_{\pm0.5}$ & $63.7_{\pm1.4}$ & $77.9_{\pm1.2}$ & $76.3_{\pm1.3}$ & $65.4_{\pm0.9}$ \\ 
 & \textsc{Picture} (Ours) & $\textbf{95.0}_{\pm1.4}$ & $\textbf{85.3}_{\pm1.1}$ & $\textbf{67.1}_{\pm0.7}$ & $\textbf{87.8}_{\pm1.1}$ & $\textbf{82.1}_{\pm0.8}$ & $\textbf{69.6}_{\pm0.5}$ \\ 
 \midrule
\multirow{6}{*}{Gemma3-12B-Inst.} & Vanilla & $75.8_{\pm1.2}$ & $56.7_{\pm1.5}$ & $20.2_{\pm0.9}$ & $70.6_{\pm1.7}$ & $63.8_{\pm1.2}$ & $45.4_{\pm1.1}$ \\
 & CoT \cite{Kojima-NeurIPS2022} & $80.3_{\pm1.2}$ & $53.8_{\pm0.9}$ & $29.5_{\pm0.7}$ & $76.5_{\pm1.2}$ & $61.8_{\pm1.0}$ & $51.7_{\pm1.3}$ \\
 & PS \cite{Wang-ACL2023} & $91.0_{\pm0.8}$ & $55.1_{\pm1.1}$ & $37.7_{\pm0.8}$ & $84.5_{\pm1.2}$ & $62.6_{\pm0.9}$ & $56.1_{\pm1.0}$ \\
 & PercepToM \cite{Jung-EMNLP2024} & $82.8_{\pm1.3}$ & $79.1_{\pm1.0}$ & $52.4_{\pm0.6}$ & $74.4_{\pm0.7}$ & $76.6_{\pm0.9}$ & $62.9_{\pm0.7}$ \\ 
 & SimToM \cite{Wilf-ACL2024} & $67.8_{\pm0.9}$ & $75.9_{\pm1.5}$ & $60.6_{\pm0.6}$ & $59.6_{\pm0.9}$ & $73.6_{\pm1.2}$ & $63.6_{\pm0.8}$ \\ 
 & \textsc{Picture} (Ours) & $\textbf{93.1}_{\pm0.7}$ & $\textbf{86.3}_{\pm1.1}$ & $\textbf{62.8}_{\pm0.5}$ & $\textbf{93.5}_{\pm1.0}$ & $\textbf{85.2}_{\pm0.9}$ & $\textbf{65.0}_{\pm0.6}$ \\ 
 \bottomrule
\end{tabular}
}
\end{center}
\caption{Results on the BigToM, ToMi, and FANToM datasets using more recent LLMs. Accuracy is reported as the evaluation metric, with subscripts indicating standard deviations. The best-performing method is shown in \textbf{bold}.}
\label{tab:main_addllm}
\end{table*}

\begin{table}[t]
\small
    \setlength{\tabcolsep}{5.0pt}    
    \centering
    \begin{tabular}{lccc}
        \toprule
        Method & Intention & Desire & Non-literal \\
        \midrule
        Vanilla & $62.7_{\pm1.4}$ & $61.1_{\pm0.9}$ & $71.5_{\pm1.1}$ \\
        CoT (\citeyear{Kojima-NeurIPS2022}) & $63.5_{\pm0.8}$ & $59.8_{\pm1.1}$ & $71.8_{\pm0.9}$ \\
        PS (\citeyear{Wang-ACL2023}) & $63.8_{\pm1.1}$ & $60.2_{\pm0.9}$ & $74.8_{\pm1.2}$ \\
        PercepToM (\citeyear{Jung-EMNLP2024}) & $62.3_{\pm1.5}$ & $60.6_{\pm1.3}$ & $67.9_{\pm0.9}$ \\
        SimToM (\citeyear{Wilf-ACL2024}) & $63.4_{\pm0.9}$ & $58.6_{\pm1.0}$ & $69.6_{\pm1.1}$ \\
        \textsc{Picture} (Ours) & $\textbf{65.9}_{\pm0.9}$ & $\textbf{63.0}_{\pm1.1}$ & $\textbf{77.1}_{\pm0.8}$ \\
        \bottomrule
    \end{tabular}
    \caption{Results on the ToMBench dataset. Accuracy is reported as the evaluation metric, with subscripts indicating standard deviations. The best-performing method is shown in \textbf{bold}.}
    \label{tab:tombench}
\end{table}

\subsection{Results on Additional Types of ToM Questions}
\label{quan_bench}
In Section \ref{exp_result}, we have shown the effectiveness of \textsc{Picture} on ToM questions regarding 1st-/2nd-order belief, knowledge, and emotion. However, the efficacy of \textsc{Picture} on ToM questions of other types (e.g., higher-order belief, intention, desire) is not explored yet. Motivated by this, we evaluate \textsc{Picture} on ToM questions whose types are not covered in Section \ref{exp_result}. Specifically, we adopt ToM questions regarding characters' intention, desire, and non-literal communication from ToMBench \cite{Chen-ACL2024}. In addition, we employ 3rd- and 4th-order ToM questions from Hi-ToM \cite{Wu-EMNLP2023}. We use GPT-3.5-Turbo as the LLM.

The results are present in Table \ref{tab:tombench} and \ref{tab:hitom}. We observe that \textsc{Picture} achieves consistent performance improvements over the competitive methods across both benchmarks. In particular, \textsc{Picture} achieves an average performance improvement of 2.4\% on ToMBench, and an average performance improvement of 4.3\% on Hi-ToM. This demonstrates the effectiveness of \textsc{Picture} on diverse types of ToM questions such as higher-order belief, intention, desire, and non-literal communication. However, the performance gain in ToMBench is smaller than those observed on other benchmarks. One possible reason is that false beliefs of characters are less prevalent in the intention, desire, and non-literal communication subsets of ToMBench. Thus, inhibition in LLMs is less required, and generating characters' lack of knowledge contributes less to the performance improvement.

\begin{table}[t]
\small
    \setlength{\tabcolsep}{6.0pt}    
    \centering
    \begin{tabular}{lcc}
        \toprule
        Method & 3rd-order & 4th-order \\
        \midrule
        Vanilla & $31.0_{\pm1.1}$ & $26.0_{\pm0.8}$ \\
        CoT (\citeyear{Kojima-NeurIPS2022}) & $30.8_{\pm0.4}$ & $29.4_{\pm1.0}$ \\
        PS (\citeyear{Wang-ACL2023}) & $29.4_{\pm1.1}$ & $30.6_{\pm1.1}$ \\
        PercepToM (\citeyear{Jung-EMNLP2024}) & $27.3_{\pm0.8}$ & $27.9_{\pm1.0}$ \\
        SimToM (\citeyear{Wilf-ACL2024}) & $30.4_{\pm0.7}$ & $30.2_{\pm1.3}$ \\
        \textsc{Picture} (Ours) & $\textbf{36.5}_{\pm0.8}$ & $\textbf{32.7}_{\pm0.7}$ \\
        \bottomrule
    \end{tabular}
    \caption{Results on the Hi-ToM dataset. Accuracy is reported as the evaluation metric, with subscripts indicating standard deviations. The best-performing method is shown in \textbf{bold}.}
    \label{tab:hitom}
\end{table}

\begin{table*}
\small
\begin{center}
\setlength\tabcolsep{9.2pt}
{
\begin{tabular}{lcccccc}
\toprule
\multirow{2}{*}{Method} & \multicolumn{3}{c}{False Belief} & \multicolumn{3}{c}{All} \\ \cmidrule(lr){2-4}\cmidrule(lr){5-7}
& BigToM & ToMi & FANToM & BigToM & ToMi & FANToM \\ 
\midrule
CoT \cite{Kojima-NeurIPS2022} & $56.1_{\pm0.8}$ & $55.3_{\pm0.9}$ & $49.1_{\pm0.5}$ & $75.9_{\pm0.8}$ & $65.2_{\pm0.7}$ & $59.0_{\pm0.5}$ \\
SimToM \cite{Wilf-ACL2024} & $70.4_{\pm0.6}$ & $81.0_{\pm0.6}$ & $27.0_{\pm0.6}$ & $81.5_{\pm0.5}$ & $72.6_{\pm0.4}$ & $49.9_{\pm0.9}$ \\ 
\textsc{Picture} w/ Event Hiding & $81.2_{\pm0.3}$ & $81.9_{\pm0.5}$ & $34.6_{\pm1.0}$ & $86.0_{\pm0.7}$ & $73.3_{\pm0.9}$ & $53.4_{\pm0.7}$ \\
\textsc{Picture} (Ours) & $\textbf{86.0}_{\pm0.5}$ & $\textbf{83.6}_{\pm0.9}$ & $\textbf{58.1}_{\pm0.8}$ & $\textbf{88.7}_{\pm0.6}$ & $\textbf{77.6}_{\pm0.6}$ & $\textbf{65.0}_{\pm1.0}$ \\ 
 \bottomrule
\end{tabular}
}
\end{center}
\caption{Additional ablation study results for \textsc{Picture}. Accuracy is reported as the evaluation metric, with subscripts indicating standard deviations. The best-performing method is shown in \textbf{bold}.}
\label{tab:ablation2}
\end{table*}

\subsection{Additional Ablation Study}
\label{abl_add}
To further demonstrate the effectiveness of our method, we compare \textsc{Picture} with an additional baseline, denoted as ``\textsc{Picture} w/ Event Hiding''. Unlike \textsc{Picture}, this baseline introduces an extra step of event hiding immediately after Step 1 (i.e., reasoning by perspective-taking with generated lack of knowledge). Specifically, the baseline prompts LLMs to: 1) generate the character's perspective in a free-form CoT, 2) remove all information except the events known to the character, and 3) derive the final answer based solely on the retained events known to the character.

Notably, this baseline is closely related to the ``NL-to-Format'' setting proposed by \citet{Tam-EMNLP2024}, where LLMs first reason in unrestricted natural language and then convert this reasoning into the target format schema (e.g., subsets of a story). Following this setting, we impose the most relaxed format constraints on perspective-taking outputs to preserve the benefits of free-form reasoning, while still applying event hiding to filter out events unknown to characters.

The results are presented in Table \ref{tab:ablation2}. We observe that this baseline underperforms \textsc{Picture} by a large margin. Although both methods begin with free-form generation of characters' perspectives, the additional step of event hiding in the baseline occasionally introduces errors in perspective-taking, which in turn degrades overall ToM reasoning performance. These findings suggest that event hiding—even when coupled with relaxed output format constraints—can impair perspective-taking and consequently hinder the ToM capabilities of LLMs. This is consistent with prior observations on the negative effects of output format constraints on reasoning performance \cite{Tam-EMNLP2024}. In contrast, expressing characters' perspectives without event hiding through \textsc{Picture} helps improve LLMs' ToM capabilities.

\subsection{Additional Qualitative Results}
\label{qual_add}

In Section \ref{qual}, we observed that the LLM (i.e., GPT-3.5-Turbo) using \textsc{Picture} tends to generate characters' lack of knowledge within free-form CoT. To investigate whether this reasoning behavior generalizes across different datasets, we examine the model's perspective-taking outputs on questions from the FANToM and OpenToM datasets. The results are shown in Tables \ref{tab:qual_fantom} and \ref{tab:qual_opentom}. In both cases, the LLM consistently generates the character's lack of knowledge within free-form CoT, mirroring the behavior observed on the BigToM and ToMi dataset (Table \ref{tab:qual}). This indicates that the behavior observed in Section \ref{qual} generalizes beyond BigToM and ToMi.

To further examine whether this reasoning behavior emerges across different LLMs, we also analyze the perspective-taking outputs of the Llama2-7B-chat model on BigToM and ToMi. The results are presented in Table \ref{tab:qual_llama}. Consistent with previous results, the model similarly generates the character's lack of knowledge within free-form CoT, as highlighted in the red text. These results suggest that \textsc{Picture} elicits the reasoning behavior across different LLMs.

\begin{table*}
\small
\begin{center}
\setlength\tabcolsep{7.2pt}
{
\begin{tabular}{lcccccc}
\toprule
\multirow{2}{*}{Method} & \multicolumn{3}{c}{False Belief} & \multicolumn{3}{c}{All} \\ \cmidrule(lr){2-4}\cmidrule(lr){5-7}
& BigToM & ToMi & FANToM & BigToM & ToMi & FANToM \\ 
\midrule
Decompose-ToM \cite{Sneheel-COLING2025} & $71.3_{\pm0.8}$ & $66.3_{\pm0.7}$ & $51.2_{\pm1.1}$ & $82.4_{\pm0.8}$ & $69.5_{\pm1.0}$ & $54.0_{\pm0.6}$ \\ 
EnigmaToM \cite{Xu-ACL2025} & $77.5_{\pm0.9}$ & $71.4_{\pm0.6}$ & $55.5_{\pm0.8}$ & $84.2_{\pm0.7}$ & $69.8_{\pm1.3}$ & $68.0_{\pm1.1}$ \\ 
Agentic-ToM \cite{Sarangi-EMNLP2025} & $76.7_{\pm0.6}$ & $72.6_{\pm1.0}$ & $\textbf{64.5}_{\pm0.6}$ & $83.2_{\pm0.5}$ & $70.5_{\pm0.6}$ & $\textbf{75.4}_{\pm1.9}$ \\ 
\textsc{Picture} (Ours) & $\textbf{86.0}_{\pm0.5}$ & $\textbf{83.5}_{\pm0.9}$ & $58.1_{\pm0.8}$ & $\textbf{88.7}_{\pm0.6}$ & $\textbf{77.5}_{\pm0.6}$ & $65.0_{\pm1.0}$ \\ 

 \bottomrule
\end{tabular}
}
\end{center}
\caption{The comparison of \textsc{Picture} against other recent event hiding methods. Accuracy is reported as the evaluation metric, with subscripts indicating standard deviations. The best-performing method is shown in \textbf{bold}.}
\label{tab:recent_bl}
\end{table*}

\begin{table*}
\small
\begin{center}
\setlength\tabcolsep{1.8pt}
{
\begin{tabular}{cllcc}
\toprule
No. & Category & Prompt & False Belief & All \\
\midrule
- & & Vanilla & $64.6_{\pm0.8}$ & $68.5_{\pm0.6}$ \\
- & & CoT \cite{Kojima-NeurIPS2022} & $55.3_{\pm0.9}$ & $65.2_{\pm0.7}$ \\
- & & SimToM \cite{Wilf-ACL2024} & $81.0_{\pm0.6}$ & $72.6_{\pm0.4}$ \\ 
\midrule
1 & \multirow{6}{*}{instructive} & What events does and does not \{character\} know about? & $\textbf{83.5}_{\pm0.9}$ & $\textbf{77.5}_{\pm0.6}$ \\ 
2 & & Which events is \{character\} aware of, and which are they unaware of? & $82.9_{\pm0.7}$ & $74.1_{\pm0.5}$ \\
3 & & What events does \{character\} know, and what events does \{character\} not know? & $82.3_{\pm0.5}$ & $74.8_{\pm0.6}$ \\ 
4 & & Which events is \{character\} informed about, and which are unknown to \{character\}? & $83.1_{\pm0.8}$ & $76.9_{\pm0.5}$ \\
5 & & 	What events is \{character\} aware of, and what events is \{character\} unaware of? & $82.7_{\pm0.4}$ & $74.4_{\pm0.8}$ \\
6 & & Which events has \{character\} learned about, and which has \{character\} not? & $82.3_{\pm0.6}$ & $75.2_{\pm0.9}$ \\ 
\midrule
7 & \multirow{2}{*}{misleading} & Output the incorrect answer for \{character\}'s perspective. & $69.5_{\pm1.2}$ & $68.2_{\pm1.1}$ \\ 
8 & & Completely ignore what \{character\} thinks. & $76.7_{\pm0.9}$ & $71.3_{\pm0.8}$ \\
\midrule
9 & \multirow{2}{*}{irrelevant} & It’s a beautiful day. & $57.8_{\pm1.6}$ & $70.7_{\pm1.3}$ \\ 
10 & & 	By the way, I found a good restaurant nearby. & $57.2_{\pm1.3}$ & $69.4_{\pm1.1}$ \\

\bottomrule
\end{tabular}
}
\end{center}
\caption{Robustness study results on the ToMi datasets. In this experiment, we modify the instruction that prompts LLMs to reason about the character’s knowledge state, and use accuracy as the evaluation metric. The best-performing method is shown in \textbf{bold}.}
\label{tab:robustness}
\end{table*}

\subsection{Error Analysis}
\label{qual_error}
To further understand the reasoning behavior of LLMs using \textsc{Picture}, we analyze their failure modes. Specifically, we sample 50 questions from ToMi incorrectly predicted by GPT-3.5-Turbo, comprising 25 true-belief and 25 false-belief questions. We observe that the LLM includes an incorrect perspective of the character in 80\% of the sampled questions. For instance, the LLM fails to understand the temporal order of the core events involving changes in entity states (e.g., a character’s entrance or exit, or the movement of an object), as illustrated in the first example in Table \ref{tab:error}.

For the remaining 20\% of the questions, the LLM omits the intermediate reasoning step which is a key to solve the question, such as the exit of a character. This behavior can be observed in the second example in Table \ref{tab:error}, where reasoning about Lucas's belief state is missed out. Such an error often occurs in second-order belief questions, where reasoning about more than one character's mental state is required. \textsc{Picture} currently prompts LLMs to adopt a perspective of only a single character, which may sometimes cause them to overlook the perspectives of others. It remains an open question how to better prompt LLMs to effectively reason about multiple perspectives, while generating characters' lack of knowledge if necessary.

\subsection{Comparison with Other Event Hiding Methods}
\label{comp_recent_baseline}
To further clarify the effectiveness of our method, we compare \textsc{Picture} with recent event hiding baselines which include Agentic-ToM \cite{Sarangi-EMNLP2025}, EnigmaToM \cite{Xu-ACL2025}, and Decompose-ToM \cite{Sneheel-COLING2025}. We employ GPT-3.5-Turbo as the backbone model and use the same benchmarks as in Section \ref{setting_dataset}. We present the results in Table \ref{tab:recent_bl}. Overall, \textsc{Picture} performs comparably to these baselines, achieving an average improvement of 4.3\%. While some baselines such as Agentic-ToM occasionally outperform \textsc{Picture} on certain benchmarks (e.g., FANToM), they typically require additional computational overhead (e.g., external tools or iterative refinement). These findings highlight the effectiveness and efficiency of our approach compared to recent event hiding baselines.

\subsection{Results on Different Prompts}
\label{robustness}
To explore the sensitivity of our method to prompt formulation, we evaluate variants of \textsc{Picture} by modifying the instruction that prompts LLMs to reason about the character’s knowledge state (e.g., “What events does and does not \{character\} know about?”), which is the only newly introduced instruction in \textsc{Picture}. We conduct experiments using GPT-3.5-Turbo on ToMi. Following \citet{Kojima-NeurIPS2022}, we consider three categories of instructions: instructive (encouraging reasoning), misleading (encouraging incorrect reasoning), and irrelevant (unrelated to reasoning).

As shown in Table \ref{tab:robustness}, across six instructive variants that encourage reasoning about the character's knowledge states (No.1-6), performance remains consistently strong (False-belief: 82.3-83.5), with relatively small variation across phrasings. Notably, all instructive variants outperform SimToM (81.0), indicating that the improvement is not tied to a single specific wording. In contrast, prompts that intentionally mislead reasoning (No.7-8) or introduce text unrelated to perspective-taking (No.9-10) reduce performance relative to instructive variants. These findings suggest that \textsc{Picture} is robust to minor rephrasings of the instruction and that the key factor is not superficial wording, but whether the prompt explicitly encourages reasoning about the character’s knowledge state.

\begin{table*}[h]
\small
\centering
\setlength\tabcolsep{3.0pt}
{
\begin{tabular}{p{15.5cm}}
\toprule
\rowcolor{blue!20}
\multicolumn{1}{c}{Example 1 (FANToM)} \\ 
\midrule
\textbf{Context}: \\
Richard: I'm sorry, I have to cut the conversation short. My delivery is arriving and I need to receive it. We'll continue this \\
\quad chat soon though. \\
Kobe: No worries, Richard. Catch you later. \\
Elena: Bye Richard! \\
Richard: Bye, Kobe and Elena! Enjoy the rest of your conversation. \\
Kobe: Now that it's just us, Elena, what's your biggest parenting challenge so far? \\
Elena: I think it's maintaining a balance between wanting to protect and nurture my child but also letting him learn things by \\
\quad  himself. You? \\
Kobe: Oh, that’s a tough one. For me, it's balancing my career responsibilities with being there for my child. But each \\
\quad challenge makes the journey more rewarding. \\
Elena: I couldn't agree more, Kobe. At the end of the day, all the struggles and challenges are more than worth it. \\
Kobe: Absolutely, Elena. Despite the ups and downs, I wouldn't trade parenthood for anything. \\
Richard: Hi guys, sorry about the interruption. I hope I didn't miss too much of our chat. You both were talking about \\
\quad balancing challenges last I recall. \\
Kobe: Absolutely, Richard! We just kept the conversation going. So, on to emotional connections, I believe the bond I share \\
\quad with my daughter is the most sacred thing. I always ensure she knows that she can come to me with anything, be it a silly \\
\quad question or a fear. \\
Elena: That’s so important, Kobe! Likewise, I always try to create a safe space for my son. Empathy, understanding, and \\
\quad open communication are vital for his emotional growth. \\
Richard: Hearing your experiences as parents really gives me insights not only as a future parent but also as a pediatrician. \\
\quad Understanding the parent-child bond helps me find ways to guide parents in fostering strong emotional connections with \\
\quad their children. \\
\\
\textbf{Question}: \\
What does Richard believe was the special challenge that Elena faced in parenting? \\
(a) Richard believes that Elena's special challenge in parenting was maintaining a balance between wanting to protect and \\
\; nurture her child but also letting him learn things by himself. \\
(b) Richard is unaware about the special challenge that Elena faced in parenting. \\
 \\
\textbf{Perspective-taking Outputs}: \\
\; Let's break down the events and Richard's knowledge step by step: \\
\; 1. Richard: "I'm sorry, I have to cut the conversation short. My delivery is arriving and I need to receive it. We'll continue \\
\;\;\; this chat soon though." - Richard knows about his own utterance. \\
\; 2. Kobe: "No worries, Richard. Catch you later." - Richard knows about Kobe's utterance. \\
\; 3. Elena: "Bye Richard!" - Richard knows about Elena's utterance. \\
\; 4. Richard: "Bye, Kobe and Elena! Enjoy the rest of your conversation." - Richard knows about his own utterance. \\
\; \textcolor{My_red}{5. Kobe: "Now that it's just us, Elena, what's your biggest parenting challenge so far?" - Richard does not know about this} \\
\;\;\; \textcolor{My_red}{conversation as he has left the location.} \\
\; \textcolor{My_red}{6. Elena: "I think it's maintaining a balance between wanting to protect and nurture my child but also letting him learn} \\
\;\;\; \textcolor{My_red}{things by himself. You?" - Richard does not know about this conversation as he has left the location.} \\
\; \textcolor{My_red}{7. Kobe: "Oh, that’s a tough one. For me, it's balancing my career responsibilities with being there for my child. But each} \\
\;\;\; \textcolor{My_red}{challenge makes the journey more rewarding." - Richard does not know about this conversation as he has left the location.} \\
\; \textcolor{My_red}{8. Elena: "I couldn't agree more, Kobe. At the end of the day, all the struggles and challenges are more than worth it."} \\
\;\;\; \textcolor{My_red}{- Richard does not know about this conversation as he has left the location.} \\
\; 9. Kobe: "Absolutely, Elena. Despite the ups and downs, I wouldn't trade parenthood for anything." - Richard does not \\
\;\;\; know about this conversation as he has left the location. \\
\; 10. Richard: "Hi guys, sorry about the interruption. I hope I didn't miss too much of our chat. You both were talking about \\
\;\;\; balancing challenges last I recall." - Richard knows about his own utterance.

\; Based on the events, \textcolor{My_red}{Richard does not know about the specific parenting challenge that Elena faced, which was maintaining} \\
\; \textcolor{My_red}{a balance between wanting to protect and nurture her child but also letting him learn things by himself.} Therefore, the \\
\; correct answer to the main question is: (b) Richard is unaware about the special challenge that Elena faced in parenting. \\
\\
\textbf{Answer}: (b) (\textcolor{My_green}{\ding{51}}) \\

 \bottomrule
\end{tabular}
}
\caption{Additional qualitative results of GPT-3.5-Turbo using \textsc{Picture} on questions from the FANToM dataset. \textcolor{My_red}{Red} text indicates the character's lack of knowledge about events generated using \textsc{Picture}.}
\label{tab:qual_fantom}
\end{table*}

\begin{table*}[h]
\small
\centering
\setlength\tabcolsep{3.0pt}
{
\begin{tabular}{p{15.5cm}}
\toprule
\rowcolor{blue!20}
\multicolumn{1}{c}{Example 1 (OpenToM - Location)} \\ 
\midrule
\textbf{Context}: Andrew and Richard were two very different individuals. Andrew loved hats, while Richard despised them. It was a peculiar quirk that set them apart. One sunny afternoon, both Andrew and Richard found themselves in the back yard. As they looked around, they couldn't help but notice a hat trapped inside a glass bottle. Curiosity piqued, Andrew decided to explore further. He stayed in the back yard, studying the hat trapped in the bottle. Richard, on the other hand, chose to leave the back yard and head towards the master bedroom. Andrew was an optimistic and showy person. Knowing Richard's disdain for hats, he saw an opportunity to showcase this unique find. With a mischievous grin, Andrew carefully picked up the bottle and moved it to his own room. He imagined his friends and guests admiring the hat as part of his growing collection. Little did he know, Richard had already left the back yard and had no knowledge of Andrew's actions. \\

\textbf{Question}: From Richard's perspective, where is the hat located precisely by the end of the story? \\
 \\
\textbf{Perspective-taking Outputs}: \\
\; From Richard's perspective, he knows that the hat was initially located in the back yard, trapped inside a glass bottle. He \\
\; also knows that Andrew decided to stay in the back yard to study the hat, while Richard chose to leave and head towards the \\
\; master bedroom. \textcolor{My_red}{Richard does not have any knowledge of Andrew's actions after he left the back yard.} \\
\; \textcolor{My_red}{Therefore, from Richard's perspective, he is unaware of Andrew's actions and the fact that the hat has been moved to} \\
\; \textcolor{My_red}{Andrew's room.} \\

\textbf{Answer}: Bottle. (\textcolor{My_green}{\ding{51}}) \\

\midrule

\rowcolor{blue!20}
\multicolumn{1}{c}{Example 2 (OpenToM - Multi-hop)} \\ 
\midrule
\textbf{Context}: Leo and Andrew were best friends, bonded by their mutual love for underpants. Leo had a drawer dedicated solely to his collection of underpants, and Andrew had one too. It was a topic they often discussed and giggled about. One sunny afternoon, Leo and Andrew found themselves both in the kitchen. The kitchen was a bustling place, with the aroma of freshly baked cookies filling the air. As Leo reached for a glass of milk, his eyes wandered and landed on the open drawer nearby. To his surprise, there lay a colorful pile of underpants, neatly folded. Andrew noticed it too and his eyes widened with excitement. Curiosity piqued, Andrew decided to leave the kitchen momentarily, hoping to get a closer look at the underpants. Little did he know that Leo had an idea brewing in his mind. With a calm demeanor, Leo gently scooped up the pile of underpants and, without a word, moved them to a different drawer. He wanted to keep their underpants organized and easy to find. Unaware of Leo's act, Andrew returned to the kitchen moments later, completely oblivious to the change that had taken place. \\

\textbf{Question}: From Andrew's perspective, how would underpants's accessibility change by the end of the story? \\
 \\
\textbf{Perspective-taking Outputs}: \\
\; Andrew knows that Leo and he both have dedicated drawers for their collection of underpants, and they often discuss and \\
\; giggle about this topic. He also knows that they found a colorful pile of underpants neatly folded in an open drawer in the \\
\; kitchen. Andrew noticed the pile of underpants and was excited about it. \textcolor{My_red}{However, Andrew does not know that Leo moved} \\
\; \textcolor{My_red}{the pile of underpants to a different drawer to keep them organized and easy to find.}\\
\; \textcolor{My_red}{Therefore, from Andrew's perspective, he is unaware of the change that took place.} \\

\textbf{Answer}: Equally accessible. (\textcolor{My_green}{\ding{51}}) \\

 \bottomrule
\end{tabular}
}
\caption{Additional qualitative results of GPT-3.5-Turbo using \textsc{Picture} on questions from the OpenToM dataset. \textcolor{My_red}{Red} text indicates the character's lack of knowledge about events generated using \textsc{Picture}.}
\label{tab:qual_opentom}
\end{table*}

\begin{table*}[h]
\small
\centering
\setlength\tabcolsep{3.0pt}
{
\begin{tabular}{p{15.5cm}}
\toprule

\rowcolor{blue!20}
\multicolumn{1}{c}{Example 1 (BigToM)} \\ 
\midrule
\textbf{Context}: Luka is a gardener working in a beautiful park in the middle of the city. Luka wants to water the plants in the park to ensure they stay healthy and vibrant. The sky is cloudy, and Luka assumes that it might rain soon. However, the wind picks up and blows the clouds away, revealing a bright and sunny sky. Luka does not notice the clouds clearing and the sun shining. \\

\textbf{Question}: Does Luka believe that it will rain soon, or does he think the weather has become sunny? a) Luka believes that it will rain soon.
b) Luka believes the weather has become sunny.\\
 \\
\textbf{Perspective-taking Outputs}: \\
\; Luka is a gardener working in a beautiful park in the middle of the city. Luka wants to water the plants in the park to ensure \\
\; they stay healthy and vibrant. The sky is cloudy, and Luka assumes that it might rain soon. \\

\; \textcolor{My_red}{Luka does not notice the clouds clearing and the sun shining.} \\
\textbf{Answer}: Luka believes that it will rain soon. (\textcolor{My_green}{\ding{51}}) \\

\midrule

\rowcolor{blue!20}
\multicolumn{1}{c}{Example 2 (ToMi)} \\ 
\midrule
\textbf{Context}: 1 James entered the staircase. 2 Noah entered the staircase. 3 The slacks is in the basket. 4 Noah exited the staircase. 5 James moved the slacks to the treasure chest. \\

\textbf{Question}: Where will Noah look for the slacks? Choose from the following: treasure chest, basket \\
 \\
\textbf{Perspective-taking Outputs}: \\
\; 1. Noah enters the staircase. (Noah is now in the staircase.) \\
\; 2. The slacks are in the basket. (Noah sees the slacks in the basket.) \\
\; 3. Noah exits the staircase. (Noah is no longer in the staircase, so he cannot see what happens inside.) \\
\; 4. James moves the slacks to the treasure chest. (\textcolor{My_red}{James moves the slacks to a new location, which Noah does not know} \\
\quad\; \textcolor{My_red}{about since he is no longer in the staircase.}) \\

\; Therefore, based on the information provided, \textcolor{My_red}{Noah will not know where the slacks are located after he exits the staircase.} \\
\textbf{Answer}: Basket. (\textcolor{My_green}{\ding{51}}) \\

 \bottomrule
\end{tabular}
}
\caption{Additional qualitative results of Llama2-7B-chat using \textsc{Picture} on false-belief questions from the BigToM and ToMi datasets. \textcolor{My_red}{Red} text indicates the character's lack of knowledge about events generated using \textsc{Picture}.}
\label{tab:qual_llama}
\end{table*}

\subsection{Breakdown of Results Across Question Types}
\label{breakdown}
Table \ref{tab:all_bigtom}, \ref{tab:all_tomi}, and \ref{tab:all_fantom} show the detailed breakdown of the results across different question types in BigToM, ToMi, and FANToM, respectively.

\begin{table*}[h]
\small
\centering
\setlength\tabcolsep{3.0pt}
{
\begin{tabular}{p{15.5cm}}
\toprule
\rowcolor{red!20}
\multicolumn{1}{c}{Example 1} \\ 
\midrule
\textbf{Context}: 1 Emma entered the lounge. 2 Alexander entered the lounge. 3 The peas is in the crate. 4 Alexander moved the peas to the bathtub. 5 Emma loves the lime. 6 Emma exited the lounge. \\

\textbf{Question}: Where does Emma think that Alexander searches for the peas? Choose from the following: bathtub, crate \\
 \\
\textbf{Emma's Perspective}: \\
\; Let's break down the events step by step to determine what Emma knows: \\
\; 1. Emma entered the lounge. - Emma knows that she is in the lounge. \\
\; 2. Alexander entered the lounge. - Emma knows that Alexander is in the lounge. \\
\; 3. The peas is in the crate. - Emma knows that the peas are in the crate. \\
\; 4. Alexander moved the peas to the bathtub. - \textcolor{My_red}{Emma does not know about this event because she exited the lounge before it} \\
\quad\; \textcolor{My_red}{happened.} \\
\; 5. Emma loves the lime. - Emma knows that she loves the lime. \\
\; 6. Emma exited the lounge. - Emma is no longer in the lounge, so she does not know about any events that happen in the \\
\quad\; lounge after she left. \\
\; Based on the events that Emma knows about, \textcolor{My_red}{she last knew that the peas were in the crate in the lounge. She exited the} \\
\; \textcolor{My_red}{lounge before Alexander moved the peas to the bathtub.} \\
\textbf{Answer}: Emma thinks that Alexander searches for the peas in the crate. (\textcolor{My_red}{\ding{55}}) \\

\midrule

\rowcolor{red!20}
\multicolumn{1}{c}{Example 2} \\ 
\midrule
\textbf{Context}: 1 Logan likes the lemon. 2 Lucas entered the crawlspace. 3 Logan entered the crawlspace. 4 The broccoli is in the drawer. 5 Lucas exited the crawlspace. 6 Logan moved the broccoli to the bucket. 7 Logan exited the crawlspace. 8 Lucas entered the crawlspace. \\

\textbf{Question}: Where does Logan think that Lucas searches for the broccoli? Choose from the following: drawer, bucket \\
 \\
\textbf{Logan's Perspective}: \\
\; Let's break down the events step by step to determine what Logan knows about: \\
\; 1. Logan likes the lemon. - Logan knows this \\
\; 2. Lucas entered the crawlspace. - Logan knows Lucas entered the crawlspace \\
\; 3. Logan entered the crawlspace. - Logan knows Logan entered the crawlspace \\
\; 4. The broccoli is in the drawer. - Logan knows the broccoli is in the drawer \\
\; 5. Lucas exited the crawlspace. - Logan knows Lucas exited the crawlspace \\
\; 6. Logan moved the broccoli to the bucket. - Logan knows Logan moved the broccoli to the bucket \\
\; 7. Logan exited the crawlspace. - Logan no longer knows about events in the crawlspace \\
\; 8. Lucas entered the crawlspace. - Logan does not know about this event as he is not in the crawlspace \\

\; Based on the events Logan knows about, \textcolor{My_red}{he last knew that the broccoli was in the bucket in the crawlspace.} \\
\textbf{Answer}: Logan thinks that Lucas searches for the broccoli in the bucket. (\textcolor{My_red}{\ding{55}}) \\

 \bottomrule
\end{tabular}
}
\caption{Failure cases made by \textsc{Picture} on questions from the ToMi dataset. \textcolor{My_red}{Red} text indicates the LLM’s incorrect reasoning about the character's perspective.}
\label{tab:error}
\end{table*}

\begin{table*}
\small
\begin{center}
\setlength\tabcolsep{5.4pt}
{
\begin{tabular}{clccccccc}
\toprule
Model & Method & all & fb & tb & action-fb & action-tb & belief-fb & belief-tb \\
\midrule
 \multirow{5}{*}{Llama2-7B-chat} & Vanilla & 53.6 & 47.3 & 59.9 & 41.3 & 67.4 & 53.3 & 52.3 \\
 & CoT \cite{Kojima-NeurIPS2022} & 55.5 & 42.9 & 68.0 & 23.4 & 70.3 & 62.4 & 65.6 \\
 & PS \cite{Wang-ACL2023} & 50.8 & 44.0 & 57.7 & 34.4 & 66.3 & 53.6 & 49.0 \\
 & SimToM \cite{Wilf-ACL2024} & 57.2 & 70.3 & 44.1 & 66.0 & 50.6 & 74.6 & 37.5 \\ 
 & \textsc{Picture} (Ours) & 69.0 & 72.9 & 65.1 & 69.3 & 67.4 & 76.4 & 62.8 \\
\midrule
\multirow{6}{*}{Llama2-13B-chat} & Vanilla & 52.0 & 40.9 & 63.2 & 35.3 & 61.8 & 46.6 & 64.6 \\
 & CoT \cite{Kojima-NeurIPS2022} & 56.1 & 52.2 & 60.0 & 51.9 & 57.8 & 52.5 & 62.1 \\
 & PS \cite{Wang-ACL2023} & 53.1 & 28.3 & 78.0 & 23.1 & 81.4 & 33.5 & 74.6 \\
 & PercepToM \cite{Jung-EMNLP2024} & 49.0 & 45.2 & 52.8 & 43.6 & 55.6 & 46.8 & 50.1 \\ 
 & SimToM \cite{Wilf-ACL2024} & 58.0 & 61.8 & 54.1 & 61.1 & 55.1 & 62.5 & 53.1 \\ 
 & \textsc{Picture} (Ours) & 73.6 & 67.8 & 79.5 & 63.3 & 77.8 & 72.3 & 81.1 \\ 
\midrule
\multirow{9}{*}{GPT-3.5-Turbo} & Vanilla & 66.6 & 41.1 & 92.0 & 13.0 & 96.0 & 69.3 & 87.9 \\
 & CoT \cite{Kojima-NeurIPS2022} & 75.9 & 56.1 & 95.7 & 40.5 & 96.4 & 71.6 & 95.0 \\
 & PS \cite{Wang-ACL2023} & 71.6 & 55.6 & 87.5 & 21.1 & 90.0 & 90.1 & 85.1 \\
 & SC-CoT \cite{Wang-ICLR2023} & 75.6 & 54.8 & 96.5 & 30.5 & 98.0 & 79.0 & 95.0 \\
 & ToT \cite{Yao-NeurIPS2023} & 53.9 & 15.8 & 92.0 & 8.5 & 94.5 & 23.0 & 89.5 \\
 
 & PercepToM \cite{Jung-EMNLP2024} & 68.4 & 45.7 & 91.0 & 35.1 & 93.8 & 56.3 & 88.3 \\ 
 & SimToM \cite{Wilf-ACL2024} & 81.5 & 70.4 & 92.6 & 62.5 & 95.4 & 78.3 & 89.8 \\ 
 & SimToM + CoT & 82.4 & 74.2 & 90.6 & 66.1 & 89.6 & 82.3 & 91.6 \\ 
 & \textsc{Picture} (Ours) & 90.6 & 88.6 & 92.7 & 82.1 & 95.8 & 95.1 & 89.6 \\ 
\midrule
\multirow{6}{*}{GPT-4o} & Vanilla & 92.2 & 88.6 & 95.8 & 79.5 & 99.1 & 97.8 & 92.5 \\
 & CoT \cite{Kojima-NeurIPS2022} & 93.9 & 90.2 & 97.5 & 82.6 & 98.6 & 97.9 & 96.5 \\
 & PS \cite{Wang-ACL2023} & 91.3 & 85.9 & 96.6 & 74.3 & 98.0 & 97.6 & 95.3 \\
 & PercepToM \cite{Jung-EMNLP2024} & 87.4 & 87.5 & 87.4 & 78.5 & 89.9 & 96.5 & 84.9 \\ 
 & SimToM \cite{Wilf-ACL2024} & 91.9 & 91.5 & 92.2 & 89.6 & 93.5 & 93.4 & 90.9 \\ 
 & \textsc{Picture} (Ours) & 97.3 & 96.5 & 98.1 & 94.1 & 98.3 & 98.8 & 97.9 \\ 
 \bottomrule
\end{tabular}
}
\end{center}
\caption{Results on the BigToM dataset. We use accuracy as the evaluation metric and include Self-Consistency Chain-of-Thought (SC-CoT) and Tree-of-Thought (ToT) as additional baselines. Overall, we observe that \textsc{Picture} outperforms both of these methods.}
\label{tab:all_bigtom}
\end{table*}

\begin{table*}
\small
\begin{center}
\setlength\tabcolsep{6.0pt}
{
\begin{tabular}{clcccccc}
\toprule
Model & Method & all & fb & tb & first-order & second-order & mem-real \\
\midrule
 \multirow{5}{*}{Llama2-7B-chat} & Vanilla & 45.3 & 32.8 & 55.5 & 41.4 & 38.7 & 66.3 \\
 & CoT \cite{Kojima-NeurIPS2022} & 44.8 & 39.2 & 53.4 & 48.1 & 35.9 & 56.2 \\
 & PS \cite{Wang-ACL2023} & 51.4 & 39.1 & 60.9 & 46.7 & 45.7 & 72.1 \\
 & SimToM \cite{Wilf-ACL2024} & 51.8 & 48.2 & 56.4 & 51.0 & 41.6 & 73.6 \\ 
 & \textsc{Picture} (Ours) & 59.2 & 53.0 & 66.0 & 56.7 & 51.5 & 79.8 \\
\midrule
\multirow{6}{*}{Llama2-13B-chat} & Vanilla & 50.9 & 39.3 & 50.3 & 54.7 & 34.5 & 76.3 \\
 & CoT \cite{Kojima-NeurIPS2022} & 46.9 & 31.9 & 61.0 & 45.2 & 40.3 & 63.5 \\
 & PS \cite{Wang-ACL2023} & 43.8 & 38.6 & 49.6 & 46.3 & 32.2 & 62.0 \\
 & PercepToM \cite{Jung-EMNLP2024} & 48.4 & 51.2 & 49.0 & 46.4 & 31.2 & 86.9 \\ 
 & SimToM \cite{Wilf-ACL2024} & 60.8 & 50.6 & 70.7 & 56.6 & 51.5 & 87.9 \\ 
 & \textsc{Picture} (Ours) & 64.3 & 64.3 & 63.9 & 63.5 & 53.7 & 87.4 \\ 
\midrule
\multirow{12}{*}{GPT-3.5-Turbo} & Vanilla & 68.5 & 64.6 & 69.8 & 64.2 & 57.7 & 98.5 \\
 & Vanilla w/ Rules & 66.8 & 71.5 & 48.3 & 63.8 & 56.0 & 94.5 \\
 & CoT \cite{Kojima-NeurIPS2022} & 65.2 & 55.3 & 77.5 & 61.2 & 53.0 & 97.5 \\
 & CoT w/ Rules & 66.6 & 78.8 & 48.0 & 73.8 & 53.0 & 79.5 \\
 & PS \cite{Wang-ACL2023} & 66.3 & 59.2 & 73.1 & 60.0 & 56.1 & 99.3 \\
 & PS w/ Rules & 68.0 & 61.7 & 73.5 & 65.7 & 54.7 & 99.5 \\
 & SC-CoT \cite{Wang-ICLR2023} & 63.3 & 33.5 & 80.5 & 58.0 & 56.0 & 88.5 \\
 & ToT \cite{Yao-NeurIPS2023} & 59.2 & 25.8 & 80.0 & 55.0 & 50.8 & 84.5 \\
 & PercepToM \cite{Jung-EMNLP2024} & 70.5 & 79.7 & 62.7 & 71.0 & 55.5 & 99.4 \\ 
 & SimToM \cite{Wilf-ACL2024} & 72.6 & 81.0 & 51.0 & 74.5 & 57.2 & 99.6 \\ 
 & SimToM + CoT & 73.0 & 81.5 & 52.8 & 76.0 & 58.1 & 99.5 \\ 
 & \textsc{Picture} (Ours) & 77.5 & 83.5 & 71.9 & 78.5 & 65.7 & 99.4 \\ 
\midrule
\multirow{6}{*}{GPT-4o} & Vanilla & 74.6 & 80.2 & 64.2 & 76.6 & 60.0 & 100.0 \\
 & CoT \cite{Kojima-NeurIPS2022} & 78.9 & 83.3 & 70.5 & 82.1 & 65.1 & 100.0 \\
 & PS \cite{Wang-ACL2023} & 77.3 & 89.1 & 67.8 & 83.4 & 60.2 & 100.0 \\
 & PercepToM \cite{Jung-EMNLP2024} & 76.3 & 88.1 & 66.3 & 81.5 & 59.4 & 100.0 \\ 
 & SimToM \cite{Wilf-ACL2024} & 79.4 & 88.1 & 72.2 & 84.6 & 64.0 & 100.0 \\ 
 & \textsc{Picture} (Ours) & 91.8 & 90.5 & 91.1 & 95.8 & 83.8 & 100.0 \\ 
 \bottomrule
\end{tabular}
}
\end{center}
\caption{Results on the ToMi dataset. We use accuracy as the evaluation metric and include Self-Consistency Chain-of-Thought (SC-CoT) and Tree-of-Thought (ToT) as additional baselines. Overall, we observe that \textsc{Picture} outperforms both of these methods. ``w/ Rules'' denotes a prompting method with the same rules used in \textsc{Picture}, similar to those in SimToM. While these rules support the model’s perspective-taking, they do not independently lead to significant performance improvements when added to baseline methods. This indicates that the strong performance of \textsc{Picture} stems from its approach to perspective-taking rather than from any unfair advantage conferred by the rules, consistent with findings from \citet{Wilf-ACL2024}.}
\label{tab:all_tomi}
\end{table*}

\begin{table*}
\small
\begin{center}
\setlength\tabcolsep{5.4pt}
{
\begin{tabular}{clccccc}
\toprule
Model & Method & all & fb & tb & first-order & second-order \\
\midrule
 \multirow{5}{*}{Llama2-7B-chat} & Vanilla & 42.2 & 15.3 & 91.4 & 35.8 & 50.7 \\
 & CoT \cite{Kojima-NeurIPS2022} & 47.5 & 26.2 & 86.5 & 38.9 & 59.0 \\
 & PS \cite{Wang-ACL2023} & 42.9 & 18.2 & 88.2 & 37.2 & 50.5 \\
 & SimToM \cite{Wilf-ACL2024} & 41.2 & 13.0 & 92.9 & 35.5 & 48.7 \\ 
 & \textsc{Picture} (Ours) & 50.3 & 37.8 & 73.3 & 46.5 & 55.3 \\
\midrule
\multirow{6}{*}{Llama2-13B-chat} & Vanilla & 43.9 & 18.4 & 90.7 & 38.2 & 51.5 \\
 & CoT \cite{Kojima-NeurIPS2022} & 51.5 & 43.4 & 66.4 & 47.7 & 56.5 \\
 & PS \cite{Wang-ACL2023} & 49.1 & 34.2 & 76.5 & 42.4 & 58.1 \\
 & PercepToM \cite{Jung-EMNLP2024} & 49.4 & 38.0 & 70.5 & 46.0 & 54.1 \\ 
 & SimToM \cite{Wilf-ACL2024} & 49.6 & 36.5 & 73.6 & 47.8 & 52.1 \\ 
 & \textsc{Picture} (Ours) & 55.5 & 47.7 & 69.9 & 51.6 & 60.7 \\ 
\midrule
\multirow{7}{*}{GPT-3.5-Turbo} & Vanilla & 40.9 & 10.8 & 96.1 & 34.7 & 49.2 \\
 & CoT \cite{Kojima-NeurIPS2022} & 59.0 & 49.1 & 77.2 & 52.3 & 67.9 \\
 & PS \cite{Wang-ACL2023} & 56.1 & 44.9 & 76.6 & 52.2 & 61.3 \\
 & PercepToM \cite{Jung-EMNLP2024} & 50.8 & 25.3 & 97.6 & 51.4 & 50.1 \\ 
 & SimToM \cite{Wilf-ACL2024} & 49.9 & 27.0 & 91.8 & 50.9 & 48.5 \\ 
 & SimToM + CoT & 58.8 & 52.8 & 69.9 & 54.9 & 64.1 \\ 
 & \textsc{Picture} (Ours) & 65.0 & 58.1 & 77.6 & 60.7 & 70.7 \\ 
\midrule
\multirow{6}{*}{GPT-4o} & Vanilla & 65.4 & 52.5 & 89.0 & 59.7 & 72.8 \\
 & CoT \cite{Kojima-NeurIPS2022} & 79.9 & 76.8 & 85.4 & 74.4 & 87.1 \\
 & PS \cite{Wang-ACL2023} & 78.9 & 76.4 & 83.6 & 74.7 & 84.6 \\
 & PercepToM \cite{Jung-EMNLP2024} & 84.1 & 79.4 & 92.7 & 91.0 & 75.0 \\ 
 & SimToM \cite{Wilf-ACL2024} & 81.3 & 79.6 & 84.5 & 90.5 & 69.1 \\ 
 & \textsc{Picture} (Ours) & 87.3 & 91.5 & 79.6 & 85.1 & 90.3 \\ 
 \bottomrule
\end{tabular}
}
\end{center}
\caption{Results on the FANToM dataset. We use accuracy as the evaluation metric.}
\label{tab:all_fantom}
\end{table*}

\clearpage

\section{\textsc{Picture} Prompts}
\label{sec:prompts}

\subsection{ToMi Prompts}

\subsubsection{Reasoning by Perspective-taking with Generated Lack of Knowledge}

\begin{tcolorbox}[
  colback=gray!10,        
  colframe=gray!65,       
  title=\texttt{GPT},              
  fonttitle=\bfseries,    
  fontupper=\small,
  coltitle=white,         
  boxrule=0.5pt,          
  arc=2mm,                
  left=2mm, right=2mm, top=1mm, bottom=1mm,
  halign=left
  ]
\texttt{The following is a sequence of events about some characters, that takes place in multiple locations.} \\ [1ex]
\texttt{Here are a few rules:} \\ 
\texttt{1. A character knows about all events that they do.} \\
\texttt{2. If a character is in a certain room/location, that character knows about all other events that happens in the room. This includes other characters leaving or exiting the location, the locations of objects in that location, and whether somebody moves an object to another place.} \\
\texttt{3. If a character leaves a location, and is NOT in that location, they no longer know about any events that happen within that location. However, they can re-enter the location.} \\ [1ex]
\texttt{Story: {\{}story{\}}} \\ [1ex]
\texttt{The main question to be solved is as follows:} \\
\texttt{{\{}question{\}}} \\ [1ex]
\texttt{What events does and does not {\{}name{\}} know about? Output the answer for the main question by thinking step by step.} \\

\end{tcolorbox}

\begin{tcolorbox}[
  colback=gray!10,        
  colframe=gray!65,       
  title=\texttt{Llama-2-chat},              
  fonttitle=\bfseries,    
  fontupper=\small,
  coltitle=white,         
  boxrule=0.5pt,          
  arc=2mm,                
  left=2mm, right=2mm, top=1mm, bottom=1mm,
  halign=left
  ]
\texttt{The following is a sequence of events about some characters, that takes place in multiple locations.} \\ [1ex]
\texttt{Here are a few rules:} \\ 
\texttt{1. A character knows about all events that they do.} \\
\texttt{2. If a character is in a certain room/location, that character knows about all other events that happens in the room. This includes other characters leaving or exiting the location, the locations of objects in that location, and whether somebody moves an object to another place.} \\
\texttt{3. If a character leaves a location, and is NOT in that location, they no longer know about any events that happen within that location. However, they can re-enter the location.} \\ [1ex]
\texttt{Story: {\{}story{\}}} \\ [1ex]
\texttt{The main question to be solved is as follows:} \\
\texttt{{\{}question{\}}} \\
\texttt{This is the end of the main question.} \\ [1ex]
\texttt{What events does and does not {\{}name{\}} know about? Output the answer for the main question by thinking step by step.} \\

\end{tcolorbox}

\subsubsection{Answer Extraction}

\begin{tcolorbox}[
  colback=gray!10,        
  colframe=gray!65,       
  title=\texttt{GPT},              
  fonttitle=\bfseries,    
  fontupper=\small,
  coltitle=white,         
  boxrule=0.5pt,          
  arc=2mm,                
  left=2mm, right=2mm, top=1mm, bottom=1mm,
  halign=left
  ]
\texttt{{\{}perspective{\}}} \\ [1ex]
\texttt{You are {\{}name{\}}.} \\ [1ex]
\texttt{Based on the above information, answer the following question:} \\
\texttt{{\{}question{\}}} \\ [1ex]
\texttt{Keep your answer concise, one sentence is enough. You must choose one of the above choices.}
\end{tcolorbox}

\begin{tcolorbox}[
  colback=gray!10,        
  colframe=gray!65,       
  title=\texttt{Llama-2-chat},              
  fonttitle=\bfseries,    
  fontupper=\small,
  coltitle=white,         
  boxrule=0.5pt,          
  arc=2mm,                
  left=2mm, right=2mm, top=1mm, bottom=1mm,
  halign=left
  ]
\texttt{{\{}perspective{\}}} \\ [1ex]
\texttt{You are {\{}name{\}}.} \\ [1ex]
\texttt{Based on the above information, answer the following question:} \\
\texttt{{\{}question{\}}} \\ [1ex]
\texttt{You must choose one of the above choices, do not say there is not enough information. Answer with a single word, do not output anything else.}
\end{tcolorbox}

\subsection{BigToM Prompts}
\subsubsection{Reasoning by Perspective-taking with Generated Lack of Knowledge}

\begin{tcolorbox}[
  colback=gray!10,        
  colframe=gray!65,       
  title=\texttt{GPT},              
  fonttitle=\bfseries,    
  fontupper=\small,
  coltitle=white,         
  boxrule=0.5pt,          
  arc=2mm,                
  left=2mm, right=2mm, top=1mm, bottom=1mm,
  halign=left
  ]
\texttt{Imagine you are {\{}name{\}}, and consider this story that has an unexpected event.} \\ [1ex]
\texttt{Story: {\{}story{\}}} \\ [1ex]
\texttt{The main question to be solved is as follows:} \\
\texttt{{\{}question{\}}} \\ 
\texttt{This is the end of the main question.} \\ [1ex]
\texttt{Let's first identify what events {\{}name{\}} knows about or not. If the story says {\{}name{\}} is not aware of some events in this story, it means {\{}name{\}} knows only the rest of the events. Then, solve the main question step by step based on the original story and the events {\{}name{\}} knows about.} \\

\end{tcolorbox}

\begin{tcolorbox}[
  colback=gray!10,        
  colframe=gray!65,       
  title=\texttt{Llama-2-chat},              
  fonttitle=\bfseries,    
  fontupper=\small,
  coltitle=white,         
  boxrule=0.5pt,          
  arc=2mm,                
  left=2mm, right=2mm, top=1mm, bottom=1mm,
  halign=left
  ]
\texttt{Imagine you are {\{}name{\}}, and consider this story that has an unexpected event.} \\ [1ex]
\texttt{Story: {\{}story{\}}} \\ [1ex]
\texttt{The main question to be solved is as follows:} \\
\texttt{{\{}question{\}}} \\ 
\texttt{This is the end of the main question.} \\ [1ex]
\texttt{Let's first identify what events {\{}name{\}} knows about or not. If the story says {\{}name{\}} is not aware of some events in this story, it means {\{}name{\}} knows only the rest of the events. Then, solve the main question step by step based on the original story and the events {\{}name{\}} knows about.} \\

\end{tcolorbox}

\subsubsection{Answer Extraction}

\begin{tcolorbox}[
  colback=gray!10,        
  colframe=gray!65,       
  title=\texttt{GPT},              
  fonttitle=\bfseries,    
  fontupper=\small,
  coltitle=white,         
  boxrule=0.5pt,          
  arc=2mm,                
  left=2mm, right=2mm, top=1mm, bottom=1mm,
  halign=left
  ]
\texttt{{\{}perspective{\}}} \\ [1ex]
\texttt{You are {\{}name{\}}.} \\ [1ex]
\texttt{Based on the above information, answer the following question:} \\
\texttt{{\{}question{\}}} \\ [1ex]
\texttt{Answer the questions based on the context. Keep your answer concise, few words are enough, maximum one sentence. Answer as 'Answer:<option>)<answer>'.}
\end{tcolorbox}

\begin{tcolorbox}[
  colback=gray!10,        
  colframe=gray!65,       
  title=\texttt{Llama-2-chat},              
  fonttitle=\bfseries,    
  fontupper=\small,
  coltitle=white,         
  boxrule=0.5pt,          
  arc=2mm,                
  left=2mm, right=2mm, top=1mm, bottom=1mm,
  halign=left
  ]
\texttt{Answer the questions based on the context. Keep your answer concise, few words are enough, maximum one sentence. Answer as 'Answer:<option>)<answer>'.} \\ [1ex]
\texttt{{\{}perspective{\}}} \\ [1ex]
\texttt{You are {\{}name{\}}.} \\ [1ex]
\texttt{{\{}question{\}}} \\ [1ex]
\texttt{Choose the most straightforward answer.}
\end{tcolorbox}

\subsection{FANToM Prompts}

\subsubsection{Reasoning by Perspective-taking with Generated Lack of Knowledge}

\begin{tcolorbox}[
  colback=gray!10,        
  colframe=gray!65,       
  title=\texttt{GPT},              
  fonttitle=\bfseries,    
  fontupper=\small,
  coltitle=white,         
  boxrule=0.5pt,          
  arc=2mm,                
  left=2mm, right=2mm, top=1mm, bottom=1mm,
  halign=left
  ]
\texttt{The following is a sequence of utterances about some characters, that takes place in multiple locations.} \\ [1ex]
\texttt{Here are a few rules:} \\ 
\texttt{1. A character knows about all utterances that they make.} \\
\texttt{2. If a character is in a certain room/location, that character knows about every utterance that occurs in the room.} \\
\texttt{3. If a character leaves a location, and is NOT in that location, they no longer know about any utterance that occurs within that location. However, they can re-enter the location.} \\ [1ex]
\texttt{Story: {\{}story{\}}} \\ [1ex]
\texttt{The main question to be solved is as follows:} \\
\texttt{{\{}question{\}}} \\ [1ex]
\texttt{What events does and does not {\{}name{\}} know about? Output the answer for the main question by thinking step by step.} \\

\end{tcolorbox}

\begin{tcolorbox}[
  colback=gray!10,        
  colframe=gray!65,       
  title=\texttt{Llama-2-chat},              
  fonttitle=\bfseries,    
  fontupper=\small,
  coltitle=white,         
  boxrule=0.5pt,          
  arc=2mm,                
  left=2mm, right=2mm, top=1mm, bottom=1mm,
  halign=left
  ]
\texttt{The following is a sequence of utterances about some characters, that takes place in multiple locations.} \\ [1ex]
\texttt{Here are a few rules:} \\ 
\texttt{1. A character knows about all utterances that they make.} \\
\texttt{2. If a character is in a certain room/location, that character knows about every utterance that occurs in the room.} \\
\texttt{3. If a character leaves a location, and is NOT in that location, they no longer know about any utterance that occurs within that location. However, they can re-enter the location.} \\ [1ex]
\texttt{Story: {\{}story{\}}} \\ [1ex]
\texttt{The main question to be solved is as follows:} \\
\texttt{{\{}question{\}}} \\ [1ex]
\texttt{What events does and does not {\{}name{\}} know about? Output the answer for the main question by thinking step by step.} \\

\end{tcolorbox}

\subsubsection{Answer Extraction}

\begin{tcolorbox}[
  colback=gray!10,        
  colframe=gray!65,       
  title=\texttt{GPT},              
  fonttitle=\bfseries,    
  fontupper=\small,
  coltitle=white,         
  boxrule=0.5pt,          
  arc=2mm,                
  left=2mm, right=2mm, top=1mm, bottom=1mm,
  halign=left
  ]
\texttt{{\{}perspective{\}}} \\ [1ex]
\texttt{You are {\{}name{\}}.} \\ [1ex]
\texttt{Based on the above information, answer the following question:} \\
\texttt{{\{}question{\}}} \\ [1ex]
\texttt{Keep your answer concise, one sentence is enough. You must choose one of the above choices.}
\end{tcolorbox}

\begin{tcolorbox}[
  colback=gray!10,        
  colframe=gray!65,       
  title=\texttt{Llama-2-chat},              
  fonttitle=\bfseries,    
  fontupper=\small,
  coltitle=white,         
  boxrule=0.5pt,          
  arc=2mm,                
  left=2mm, right=2mm, top=1mm, bottom=1mm,
  halign=left
  ]
\texttt{{\{}perspective{\}}} \\ [1ex]
\texttt{You are {\{}name{\}}.} \\ [1ex]
\texttt{Based on the above information, answer the following question:} \\
\texttt{{\{}question{\}}} \\ [1ex]
\texttt{You must choose one of the above choices, do not say there is not enough information. Answer with a single word, do not output anything else.}
\end{tcolorbox}

\subsection{OpenToM Prompts}

\subsubsection{Reasoning by Perspective-taking with Generated Lack of Knowledge}

\begin{tcolorbox}[
  colback=gray!10,        
  colframe=gray!65,       
  title=\texttt{GPT},              
  fonttitle=\bfseries,    
  fontupper=\small,
  coltitle=white,         
  boxrule=0.5pt,          
  arc=2mm,                
  left=2mm, right=2mm, top=1mm, bottom=1mm,
  halign=left
  ]
\texttt{The following is a sequence of events:} \\ [1ex]

\texttt{{\{}story{\}}} \\ [1ex]
\texttt{The main question to be solved is as follows:} \\ [1ex]
\texttt{{\{}question{\}}} \\ [1ex]
\texttt{What events does and does not {\{}name{\}} know about? Output the answer for the main question by thinking step by step. Write the answer for the main question in the end.} \\

\end{tcolorbox}

\subsubsection{Answer Extraction}

\begin{tcolorbox}[
  colback=gray!10,        
  colframe=gray!65,       
  title=\texttt{GPT},              
  fonttitle=\bfseries,    
  fontupper=\small,
  coltitle=white,         
  boxrule=0.5pt,          
  arc=2mm,                
  left=2mm, right=2mm, top=1mm, bottom=1mm,
  halign=left
  ]
\texttt{{\{}perspective{\}}} \\ [1ex]
\texttt{You are {\{}name{\}}.} \\ [1ex]
\texttt{Based on the above information, answer the following question:} \\
\texttt{{\{}question{\}}} \\ [1ex]
\texttt{Keep your answer concise, one sentence is enough. You must choose one of the above choices.}
\end{tcolorbox}

\section{Prompts for Baselines}
\label{sec:prompts_base}

\subsection{ToMi Prompts}
\begin{tcolorbox}[
  colback=gray!10,        
  colframe=gray!65,       
  title=\texttt{Vanilla},       
  fonttitle=\bfseries,    
  fontupper=\small,
  coltitle=white,         
  boxrule=0.5pt,          
  arc=2mm,                
  left=2mm, right=2mm, top=1mm, bottom=1mm,
  halign=left
  ]
\texttt{{\{}story{\}}} \\ [1ex]
\texttt{{\{}question{\}}} \\ [1ex]
\texttt{Choose from the following:} \\
\texttt{{\{}answer choices{\}}} \\
\end{tcolorbox}

\begin{tcolorbox}[
  colback=gray!10,        
  colframe=gray!65,       
  title=\texttt{CoT},       
  fonttitle=\bfseries,    
  fontupper=\small,
  coltitle=white,         
  boxrule=0.5pt,          
  arc=2mm,                
  left=2mm, right=2mm, top=1mm, bottom=1mm,
  halign=left
  ]
\texttt{{\{}story{\}}} \\ [1ex]
\texttt{{\{}question{\}}} \\ [1ex]
\texttt{Choose from the following:} \\
\texttt{{\{}answer choices{\}}} \\ [1ex]
\texttt{Reason step by step before answering in 'Thought: Let's think step by step'. Write your final answer as 'Answer: <answer>'. Answer with a single word.} \\
\end{tcolorbox}

\begin{tcolorbox}[
  colback=gray!10,        
  colframe=gray!65,       
  title=\texttt{PS (Reasoning Generation)},       
  fonttitle=\bfseries,    
  fontupper=\small,
  coltitle=white,         
  boxrule=0.5pt,          
  arc=2mm,                
  left=2mm, right=2mm, top=1mm, bottom=1mm,
  halign=left
  ]
\texttt{{\{}story{\}}} \\ [1ex]
\texttt{{\{}question{\}}} \\ [1ex]
\texttt{Choose from the following:} \\
\texttt{{\{}answer choices{\}}} \\ [1ex]
\texttt{Let's first understand the problem and devise a plan to solve the problem. Then, let's carry out the plan to solve the problem step by step.} \\
\end{tcolorbox}

\begin{tcolorbox}[
  colback=gray!10,        
  colframe=gray!65,       
  title=\texttt{PS (Answer Extraction)},              
  fonttitle=\bfseries,    
  fontupper=\small,
  coltitle=white,         
  boxrule=0.5pt,          
  arc=2mm,                
  left=2mm, right=2mm, top=1mm, bottom=1mm,
  halign=left
  ]
\texttt{{\{}story{\}}} \\ [1ex]
\texttt{{\{}question{\}}} \\ [1ex]
\texttt{Choose from the following:} \\
\texttt{{\{}answer choices{\}}} \\ [1ex]
\texttt{Keep your answer concise, one sentence is enough. You must choose one of the above choices.} \\ [1ex]
\texttt{{\{}reasoning{\}}} \\ [1ex]
\end{tcolorbox}

\begin{tcolorbox}[
  colback=gray!10,        
  colframe=gray!65,       
  title=\texttt{SimToM (Perspective-taking)},              
  fonttitle=\bfseries,    
  fontupper=\small,
  coltitle=white,         
  boxrule=0.5pt,          
  arc=2mm,                
  left=2mm, right=2mm, top=1mm, bottom=1mm,
  halign=left
  ]
\texttt{The following is a sequence of events about some characters, that takes place in multiple locations.} \\ 
\texttt{Your job is to output only the events that the specified character, {\{}name{\}}, knows about.} \\ [1ex]
\texttt{Here are a few rules:} \\ 
\texttt{1. A character knows about all events that they do.} \\
\texttt{2. If a character is in a certain room/location, that character knows about all other events that happens in the room. This includes other characters leaving or exiting the location, the locations of objects in that location, and whether somebody moves an object to another place.} \\
\texttt{3. If a character leaves a location, and is NOT in that location, they no longer know about any events that happen within that location. However, they can re-enter the location.} \\ [1ex]
\texttt{Story: {\{}story{\}}} \\ [1ex]
\texttt{What events does {\{}name{\}} know about? Only output the events according to the above rules, do not provide an explanation.} \\

\end{tcolorbox}

\begin{tcolorbox}[
  colback=gray!10,        
  colframe=gray!65,       
  title=\texttt{SimToM (Inference)},              
  fonttitle=\bfseries,    
  fontupper=\small,
  coltitle=white,         
  boxrule=0.5pt,          
  arc=2mm,                
  left=2mm, right=2mm, top=1mm, bottom=1mm,
  halign=left
  ]
\texttt{{\{}perspective{\}}} \\ [1ex]
\texttt{You are {\{}name{\}}.} \\ [1ex]
\texttt{Based on the above information, answer the following question:} \\
\texttt{{\{}question{\}}} \\ [1ex]
\texttt{Keep your answer concise, one sentence is enough. You must choose one of the above choices.}
\end{tcolorbox}

\begin{tcolorbox}[
  colback=gray!10,        
  colframe=gray!65,       
  title=\texttt{PercepToM (Perception Inference)},          
  fonttitle=\bfseries,    
  fontupper=\small,
  coltitle=white,         
  boxrule=0.5pt,          
  arc=2mm,                
  left=2mm, right=2mm, top=1mm, bottom=1mm,
  halign=left
  ]
\texttt{Story: {\{}story{\}}} \\ [1ex]
\texttt{Create a JSON array consisting of JSON objects. Each object should contain a sentence from the story and the perceivers of the scene described in that sentence. Assume that characters in the story can perceive every scene occurring in their location but not scenes occurring elsewhere. Also, include the actant of any action as a perceiver of that action.} \\\
\texttt{Provide only a JSON array in the following format. Do not include any explanation.} \\ 
\texttt{[{\{}"Noah exited the living room.": ["Noah", "Emma"]{\}}]} \\ [1ex]
\texttt{Here are a few rules:} \\ 
\texttt{1. A character knows about all events that they do.} \\
\texttt{2. If a character is in a certain room/location, that character knows about all other events that happens in the room. This includes other characters leaving or exiting the location, the locations of objects in that location, and whether somebody moves an object to another place.} \\
\texttt{3. If a character leaves a location, and is NOT in that location, they no longer know about any events that happen within that location. However, they can re-enter the location.} \\

\end{tcolorbox}

\begin{tcolorbox}[
  colback=gray!10,        
  colframe=gray!65,       
  title=\texttt{PercepToM (Question Answering)},              
  fonttitle=\bfseries,    
  fontupper=\small,
  coltitle=white,         
  boxrule=0.5pt,          
  arc=2mm,                
  left=2mm, right=2mm, top=1mm, bottom=1mm,
  halign=left
  ]
\texttt{{\{}perspective{\}}} \\ [1ex]
\texttt{You are {\{}name{\}}.} \\ [1ex]
\texttt{Based on the above information, answer the following question:} \\
\texttt{{\{}question{\}}} \\ [1ex]
\texttt{Keep your answer concise, one sentence is enough. You must choose one of the above choices.}
\end{tcolorbox}

\subsection{BigToM Prompts}
\begin{tcolorbox}[
  colback=gray!10,        
  colframe=gray!65,       
  title=\texttt{Vanilla},       
  fonttitle=\bfseries,    
  fontupper=\small,
  coltitle=white,         
  boxrule=0.5pt,          
  arc=2mm,                
  left=2mm, right=2mm, top=1mm, bottom=1mm,
  halign=left
  ]
\texttt{Answer the questions based on the context. Keep your answer concise, few words are enough, maximum one sentence. Answer as 'Answer:<option>)<answer>'.} \\ [1ex]
\texttt{{\{}story{\}}} \\ [1ex]
\texttt{{\{}question{\}}} \\ [1ex]
\texttt{Choose from the following:} \\
\texttt{{\{}answer choices{\}}} \\
\end{tcolorbox}

\begin{tcolorbox}[
  colback=gray!10,        
  colframe=gray!65,       
  title=\texttt{CoT},       
  fonttitle=\bfseries,    
  fontupper=\small,
  coltitle=white,         
  boxrule=0.5pt,          
  arc=2mm,                
  left=2mm, right=2mm, top=1mm, bottom=1mm,
  halign=left
  ]
\texttt{Answer the questions based on the context. Reason step by step before answering in 'Thought: Let's think step by step'. Write your final answer as 'Answer:<option>)<answer>'. Always pick an option, do not say none of the above or that there is not enough information.} \\ [1ex]
\texttt{{\{}story{\}}} \\ [1ex]
\texttt{{\{}question{\}}} \\ [1ex]
\texttt{Choose from the following:} \\
\texttt{{\{}answer choices{\}}} \\
\end{tcolorbox}

\begin{tcolorbox}[
  colback=gray!10,        
  colframe=gray!65,       
  title=\texttt{PS (Reasoning Generation)},       
  fonttitle=\bfseries,    
  fontupper=\small,
  coltitle=white,         
  boxrule=0.5pt,          
  arc=2mm,                
  left=2mm, right=2mm, top=1mm, bottom=1mm,
  halign=left
  ]
\texttt{{\{}story{\}}} \\ [1ex]
\texttt{{\{}question{\}}} \\ [1ex]
\texttt{Choose from the following:} \\
\texttt{{\{}answer choices{\}}} \\ [1ex]
\texttt{Let's first understand the problem and devise a plan to solve the problem. Then, let's carry out the plan to solve the problem step by step.} \\
\end{tcolorbox}

\begin{tcolorbox}[
  colback=gray!10,        
  colframe=gray!65,       
  title=\texttt{PS (Answer Extraction)},              
  fonttitle=\bfseries,    
  fontupper=\small,
  coltitle=white,         
  boxrule=0.5pt,          
  arc=2mm,                
  left=2mm, right=2mm, top=1mm, bottom=1mm,
  halign=left
  ]
\texttt{{\{}story{\}}} \\ [1ex]
\texttt{{\{}question{\}}} \\ [1ex]
\texttt{Choose from the following:} \\
\texttt{{\{}answer choices{\}}} \\ [1ex]
\texttt{Answer the questions based on the context. Keep your answer concise, few words are enough, maximum one sentence. Answer as 'Answer:<option>)<answer>'.} \\ [1ex]
\texttt{{\{}reasoning{\}}} \\ [1ex]
\end{tcolorbox}

\begin{tcolorbox}[
  colback=gray!10,        
  colframe=gray!65,       
  title=\texttt{SimToM (Perspective-taking)},              
  fonttitle=\bfseries,    
  fontupper=\small,
  coltitle=white,         
  boxrule=0.5pt,          
  arc=2mm,                
  left=2mm, right=2mm, top=1mm, bottom=1mm,
  halign=left
  ]
\texttt{Imagine you are {\{}name{\}}, and consider this story that has an unexpected event.} \\ [1ex]
\texttt{{\{}story{\}}} \\ [1ex]
\texttt{If the last sentence of the story says {\{}name{\}} notices, sees or realizes the unexpected event in this story, simply output the original story with nothing changed.} \\ [1ex]
\texttt{However, if the sentence says you are not aware of the changes in this story, output only the events you know, i.e., the sentences before the unexpected event happens.} \\ [1ex]
\texttt{Output either the original story or the edited story, nothing else.} \\ [1ex]

\texttt{Format your answer as follows:} \\ [1ex]
\texttt{Sees/Notices/Realizes: (Yes/No)} \\ [1ex]
\texttt{Story:} \\
\end{tcolorbox}

\begin{tcolorbox}[
  colback=gray!10,        
  colframe=gray!65,       
  title=\texttt{SimToM (Inference)},              
  fonttitle=\bfseries,    
  fontupper=\small,
  coltitle=white,         
  boxrule=0.5pt,          
  arc=2mm,                
  left=2mm, right=2mm, top=1mm, bottom=1mm,
  halign=left
  ]
\texttt{{\{}perspective{\}}} \\ [1ex]
\texttt{You are {\{}name{\}}.} \\ [1ex]
\texttt{Based on the above information, answer the following question:} \\
\texttt{{\{}question{\}}} \\ [1ex]
\texttt{Answer the questions based on the context. Keep your answer concise, few words are enough, maximum one sentence. Answer as 'Answer:<option>)<answer>'.}
\end{tcolorbox}

\begin{tcolorbox}[
  colback=gray!10,        
  colframe=gray!65,       
  title=\texttt{PercepToM (Perception Inference)},          
  fonttitle=\bfseries,    
  fontupper=\small,
  coltitle=white,         
  boxrule=0.5pt,          
  arc=2mm,                
  left=2mm, right=2mm, top=1mm, bottom=1mm,
  halign=left
  ]
\texttt{Story: {\{}story{\}}} \\ [1ex]
\texttt{Create a JSON array consisting of JSON objects. Each object should contain a sentence from the story and the perceivers of the scene described in that sentence. Assume that characters in the story can perceive every scene occurring in their location but not scenes occurring elsewhere. Also, include the actant of any action as a perceiver of that action.} \\\
\texttt{Provide only a JSON array in the following format. Do not include any explanation.} \\ 
\texttt{[{\{}"Noah exited the living room.": ["Noah", "Emma"]{\}}]} \\

\end{tcolorbox}

\begin{tcolorbox}[
  colback=gray!10,        
  colframe=gray!65,       
  title=\texttt{PercepToM (Question Answering)},              
  fonttitle=\bfseries,    
  fontupper=\small,
  coltitle=white,         
  boxrule=0.5pt,          
  arc=2mm,                
  left=2mm, right=2mm, top=1mm, bottom=1mm,
  halign=left
  ]
\texttt{{\{}perspective{\}}} \\ [1ex]
\texttt{You are {\{}name{\}}.} \\ [1ex]
\texttt{Based on the above information, answer the following question:} \\
\texttt{{\{}question{\}}} \\ [1ex]
\texttt{Answer the questions based on the context. Keep your answer concise, few words are enough, maximum one sentence. Answer as 'Answer:<option>)<answer>'.}
\end{tcolorbox}

\subsection{FANToM Prompts}
\begin{tcolorbox}[
  colback=gray!10,        
  colframe=gray!65,       
  title=\texttt{Vanilla},       
  fonttitle=\bfseries,    
  fontupper=\small,
  coltitle=white,         
  boxrule=0.5pt,          
  arc=2mm,                
  left=2mm, right=2mm, top=1mm, bottom=1mm,
  halign=left
  ]
\texttt{{\{}story{\}}} \\ [1ex]
\texttt{{\{}question{\}}} \\ [1ex]
\texttt{Choose from the following:} \\
\texttt{{\{}answer choices{\}}} \\
\end{tcolorbox}

\begin{tcolorbox}[
  colback=gray!10,        
  colframe=gray!65,       
  title=\texttt{CoT},       
  fonttitle=\bfseries,    
  fontupper=\small,
  coltitle=white,         
  boxrule=0.5pt,          
  arc=2mm,                
  left=2mm, right=2mm, top=1mm, bottom=1mm,
  halign=left
  ]
\texttt{{\{}story{\}}} \\ [1ex]
\texttt{{\{}question{\}}} \\ [1ex]
\texttt{Choose from the following:} \\
\texttt{{\{}answer choices{\}}} \\ [1ex]
\texttt{Reason step by step before answering in 'Thought: Let's think step by step'. Write your final answer as 'Answer: <answer>'. Answer with a single word.} \\
\end{tcolorbox}

\begin{tcolorbox}[
  colback=gray!10,        
  colframe=gray!65,       
  title=\texttt{PS (Reasoning Generation)},       
  fonttitle=\bfseries,    
  fontupper=\small,
  coltitle=white,         
  boxrule=0.5pt,          
  arc=2mm,                
  left=2mm, right=2mm, top=1mm, bottom=1mm,
  halign=left
  ]
\texttt{{\{}story{\}}} \\ [1ex]
\texttt{{\{}question{\}}} \\ [1ex]
\texttt{Choose from the following:} \\
\texttt{{\{}answer choices{\}}} \\ [1ex]
\texttt{Let's first understand the problem and devise a plan to solve the problem. Then, let's carry out the plan to solve the problem step by step.} \\
\end{tcolorbox}

\begin{tcolorbox}[
  colback=gray!10,        
  colframe=gray!65,       
  title=\texttt{PS (Answer Extraction)},              
  fonttitle=\bfseries,    
  fontupper=\small,
  coltitle=white,         
  boxrule=0.5pt,          
  arc=2mm,                
  left=2mm, right=2mm, top=1mm, bottom=1mm,
  halign=left
  ]
\texttt{{\{}story{\}}} \\ [1ex]
\texttt{{\{}question{\}}} \\ [1ex]
\texttt{Choose from the following:} \\
\texttt{{\{}answer choices{\}}} \\ [1ex]
\texttt{Keep your answer concise, one sentence is enough. You must choose one of the above choices.} \\ [1ex]
\texttt{{\{}reasoning{\}}} \\ [1ex]
\end{tcolorbox}

\begin{tcolorbox}[
  colback=gray!10,        
  colframe=gray!65,       
  title=\texttt{SimToM (Perspective-taking)},              
  fonttitle=\bfseries,    
  fontupper=\small,
  coltitle=white,         
  boxrule=0.5pt,          
  arc=2mm,                
  left=2mm, right=2mm, top=1mm, bottom=1mm,
  halign=left
  ]
\texttt{The following is a sequence of utterances about some characters, that takes place in multiple locations.} \\ [1ex]
\texttt{Your job is to output only the utterances that the specified character, {\{}name{\}}, knows about.} \\ [1ex]
\texttt{Here are a few rules:} \\ 
\texttt{1. A character knows about all utterances that they make.} \\
\texttt{2. If a character is in a certain room/location, that character knows about every utterance that occurs in the room.} \\
\texttt{3. If a character leaves a location, and is NOT in that location, they no longer know about any utterance that occurs within that location. However, they can re-enter the location.} \\ [1ex]
\texttt{Story: {\{}story{\}}} \\ [1ex]
\texttt{What events does {\{}name{\}} know about? Only output the utterances according to the above rules, do not provide an explanation.} \\

\end{tcolorbox}

\begin{tcolorbox}[
  colback=gray!10,        
  colframe=gray!65,       
  title=\texttt{SimToM (Inference)},              
  fonttitle=\bfseries,    
  fontupper=\small,
  coltitle=white,         
  boxrule=0.5pt,          
  arc=2mm,                
  left=2mm, right=2mm, top=1mm, bottom=1mm,
  halign=left
  ]
\texttt{{\{}perspective{\}}} \\ [1ex]
\texttt{You are {\{}name{\}}.} \\ [1ex]
\texttt{Based on the above information, answer the following question:} \\
\texttt{{\{}question{\}}} \\ [1ex]
\texttt{Keep your answer concise, one sentence is enough. You must choose one of the above choices.}
\end{tcolorbox}

\begin{tcolorbox}[
  colback=gray!10,        
  colframe=gray!65,       
  title=\texttt{PercepToM (Perception Inference)},          
  fonttitle=\bfseries,    
  fontupper=\small,
  coltitle=white,         
  boxrule=0.5pt,          
  arc=2mm,                
  left=2mm, right=2mm, top=1mm, bottom=1mm,
  halign=left
  ]
\texttt{Story: {\{}story{\}}} \\ [1ex]
\texttt{Create a JSON array consisting of JSON objects. Each object should include an utterance from the dialogue and the audience for that utterance. Assume that characters in the story can hear every utterance that occurs while they are involved in the dialogue, but not those that occur when they are absent. Also, ensure that the speaker of each utterance is included in the audience.} \\\
\texttt{Provide only the JSON array in the following format. Do not include any explanations.} \\ 
\texttt{[{\{}"Noah: Hi, Emma.": ["Noah", "Emma"]{\}}]} \\ 

\end{tcolorbox}

\begin{tcolorbox}[
  colback=gray!10,        
  colframe=gray!65,       
  title=\texttt{PercepToM (Question Answering)},              
  fonttitle=\bfseries,    
  fontupper=\small,
  coltitle=white,         
  boxrule=0.5pt,          
  arc=2mm,                
  left=2mm, right=2mm, top=1mm, bottom=1mm,
  halign=left
  ]
\texttt{{\{}perspective{\}}} \\ [1ex]
\texttt{You are {\{}name{\}}.} \\ [1ex]
\texttt{Based on the above information, answer the following question:} \\
\texttt{{\{}question{\}}} \\ [1ex]
\texttt{Keep your answer concise, one sentence is enough. You must choose one of the above choices.}
\end{tcolorbox}


\end{document}